\documentclass{article}

\usepackage{arxiv}

\usepackage{amsfonts}  
\usepackage{amsmath} 
\usepackage{float}
\usepackage{longtable}
\usepackage{array}
\usepackage{makecell}
\usepackage{siunitx}
\usepackage[T1]{fontenc}    
\usepackage{hyperref}       
\usepackage{url}            
\usepackage{hyperref}
\hypersetup{
    breaklinks=true,
    colorlinks=true,
    linkcolor=blue,
    urlcolor=blue,
    citecolor=blue
}

\usepackage{booktabs}       
\usepackage{amsfonts}       
\usepackage{nicefrac}       
\usepackage{microtype}      
\usepackage{cleveref}       
\usepackage{lipsum}         
\usepackage{graphicx}
\usepackage{natbib}
\usepackage{doi}
\usepackage{parskip} 

\title{A Multimedia Framework for Continuum Robots: Systematic, Computational, and Control Perspectives}

\date{}

\newif\ifuniqueAffiliation
\uniqueAffiliationtrue

\ifuniqueAffiliation 
\author{ \href{https://orcid.org/0009-0001-2682-138X}{\includegraphics[scale=0.06]{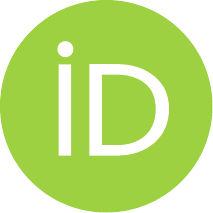}\hspace{1mm}Po-Yu Hsieh} \\
	Graduate Institute of Architecture\\
	National Yang Ming Chiao Tung University\\
	Hsinchu, Taiwan \\
	\texttt{kevinhsieh870118@arch.nycu.edu.tw} \\
	\And
	\href{https://orcid.org/0000-0002-8362-7719}{\includegraphics[scale=0.06]{orcid.pdf}\hspace{1mm}June-Hao Hou} \\
	Graduate Institute of Architecture\\
	National Yang Ming Chiao Tung University\\
	Hsinchu, Taiwan \\
	\texttt{jhou@arch.nycu.edu.tw} \\
}

\renewcommand{\headeright}{}
\renewcommand{\undertitle}{}
\renewcommand{\shorttitle}{Comparative Evaluation of Learning Models for Bionic Robots}

\hypersetup{
pdftitle={Comparative Evaluation of Learning Models for Bionic Robots: Non-Linear Transfer Function Identifications},
pdfsubject={stat.ML},
pdfauthor={Po-Yu Hsieh, June-Hao Hou},
pdfkeywords={Model-free Control, Bionic Robot, Learning-based method, Multi-input Multi-output, Transfer Function Identification},
}

\begin{document}
\maketitle

\begin{abstract}
	The control and modeling of robot dynamics have increasingly adopted model-free control strategies using machine learning.  Given the non-linear elastic nature of bionic robotic systems, learning-based methods provide reliable alternatives by utilizing numerical data to establish a direct mapping from actuation inputs to robot trajectories without complex kinematics models. However, for developers, the method of identifying an appropriate learning model for their specific bionic robots and further constructing the transfer function has not been thoroughly discussed. Thus, this research introduces a comprehensive evaluation strategy and framework for the application of model-free control, including data collection, learning model selection, comparative analysis, and transfer function identification to effectively deal with the multi-input multi-output (MIMO) robotic data.
\end{abstract}


\section{Introduction}
One of the main objectives of robotics is to develop universal robots capable of learning and adapting to varying environments \citep{relano2022modeling}. While exhibiting exceptional dexterity and precision of performing tasks in predetermined conditions, traditional rigid robots often lack the flexibility and adaptability inherently \citep{trivedi2008soft}.

\par Therefore, roboticists have drawn inspirations from nature, attempting to replicate the versatility and adaptive movements of biological systems. Biological entities maintain balance dynamically, constantly making micro-adjustments. This results in a stable yet agile locomotion that can quickly respond to disturbances and changes of current surroundings. The inherent compliance and deformable body structure of biological systems often rely on the replacement of rigid joints. Instead, flexible joints with complex tension-compression synergy, which demonstrate high degrees of freedom, are commonly observed \citep{zappetti2020variable}. This unique feature ensures versatile movements with natural grace, fluidity, and sophistication, which conventional robots with fixed joints can hardly achieve.

\par In recent decades, new designs and approaches in bionic robotics have continuously emerged, with a specific focus on effective modeling and control of biological dynamics. Despite there are various types and applications of bionic robots, the control strategies are mainly based on either a model-based or model-free method. The model-based method involves detailed mathematical descriptions and kinematic calculations to build an analytical model. Its performance largely depends on the modeling accuracy, integrity, and computational power. However, due to the elastic nature and deformable feature of bionic robots, this method is computationally intensive and robot specific \citep{vikas2015model}. The dilemma of accuracy and computation complexity has led to a paradigm shift to the model-free method \citep{yu2024biomimetic}, which can be summarized as the following:

\begin{enumerate}
    \item Data-driven: It relies heavily on collecting and utilizing data to identify the relationship between inputs (control parameters) and outputs (robot actions), which allows the robotic system to learn from experience and adapt to new situations in the long run.
    \item End-to-End mapping: This method maps the actuation inputs directly to the robot trajectory without requiring complex kinematics models, presenting less computational complexity.
    \item Utilization of learning-based models: Regarding performance optimization, various machine learning models are utilized in order to approximate an optimal control based on numerical data.
\end{enumerate}

\par Related studies \citep{giorelli2015neural} \citep{giorelli2013feed} have demonstrated that given the same inverse kinematics problem of a bionic tendon-driven robot, the model-free solution (feedforward neural network in this case) provides more accuracy and efficiency compared to the model-based analytical method (Jacobian method). In fact, ever since the early research on learning-based method for bionic robots, proposed by \citet{vikas2015model}, were introduced, the applications have been explored in various fields, such as surgical manipulators using regression models \citep{https://doi.org/10.1002/rcs.1774}, and bionic continuum robot using adaptive neural network \citep{melingui2015adaptive}. Recent studies have focused on using varying machine learning approaches to acquire an optimized transfer function due to its effectiveness of handling extensive numerical input-output data.  

\par However, since the model-free control strategy for bionic robots is a relatively new field, the method of identifying an appropriate learning model for each specific robot and constructing transfer functions have not been thoroughly discussed. The lack of an overall review and demonstration has posed uncertainties to most robot developers. Therefore, this research aims to present a comparative evaluation of learning models based on two key indexes: accuracy and computational complexity when replicating biological movements.

\section{Background and Related Works}

\subsection{Description of the bionic robot}
\begin{figure}[htbp]
    \centering
    \includegraphics[width=.9\linewidth]{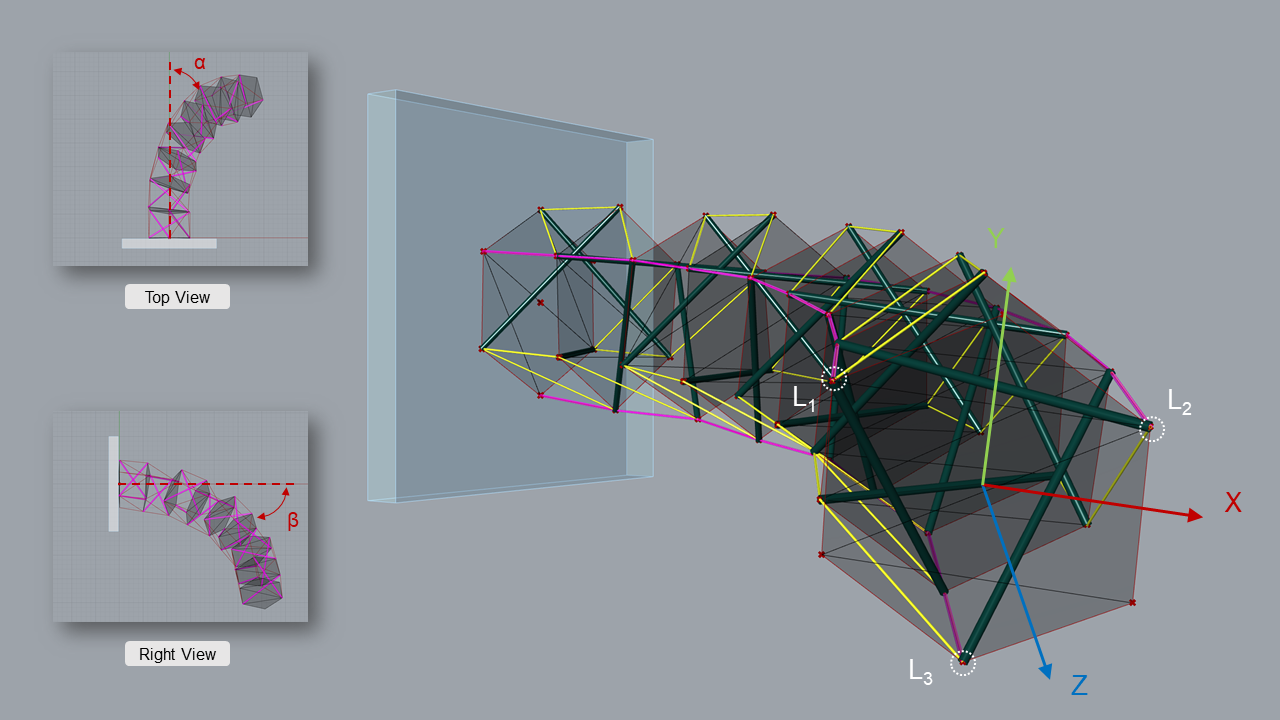} 
    \caption{View of the tendon-driven continuum robot.}
    \label{fig:intro}
\end{figure}

This research is based on a tendon-driven continuum robot with the tensegrity-based compliant mechanism (Figure \ref{fig:intro}) developed by \citet{hsiehbiomimetic}. Since much developments of robot structure and actuation system have been explored, this prototype provides a robust foundation for the research project. The actuator of this robot relies on a tendon-driven technique, grafting the parallel compliant mechanism onto a tensegrity-based robot body. Once pulling forces are applied, the robot body will gradually change its configuration to achieve the new equilibrium, which shares a similarity with biological movements. By manipulating the length variations of each tendon ($\Delta$L), the robot is able to perform a wide range of motions with inherent high compliance. Nevertheless, the modeling and control of its dynamics still require further identifications. The control method proposed in this previous paper relies on an inverse kinematics model similar to delta robots, which lacks the consideration of the non-linear elastic nature of continuum robots.

\subsection{Decoupled transfer function}

\begin{figure}[!htbp]
    \centering
    \begin{minipage}[c]{.32\textwidth}
        \centering
        \includegraphics[width=\textwidth]{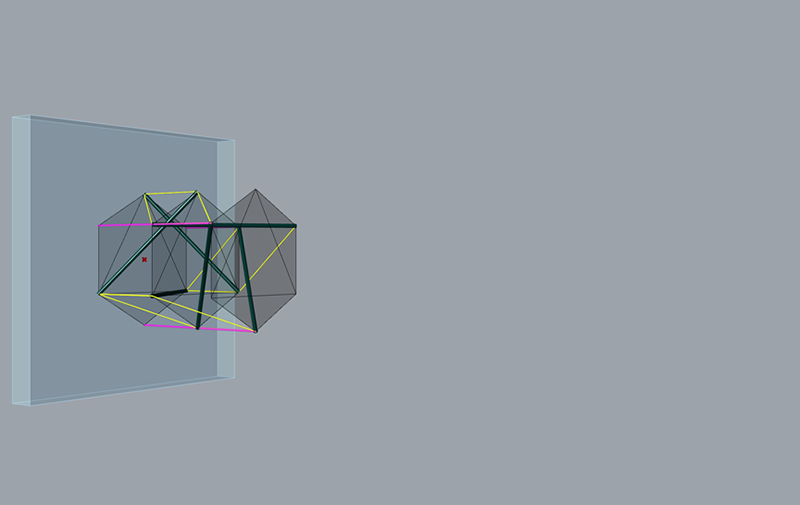}
        (a) N=1
    \end{minipage}
    \begin{minipage}[c]{.32\textwidth}
        \centering
        \includegraphics[width=\textwidth]{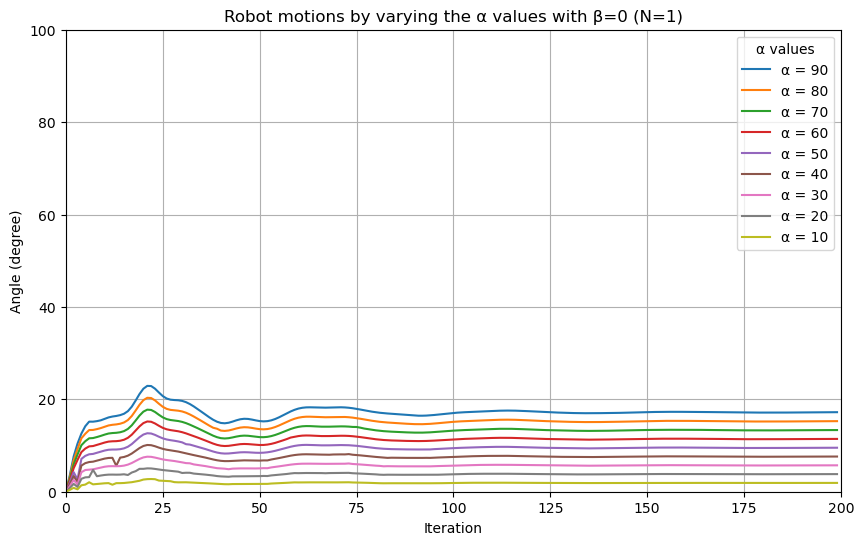}
        (b) $\alpha$ over time
    \end{minipage}
    \begin{minipage}[c]{.32\textwidth}
        \centering
        \includegraphics[width=\textwidth]{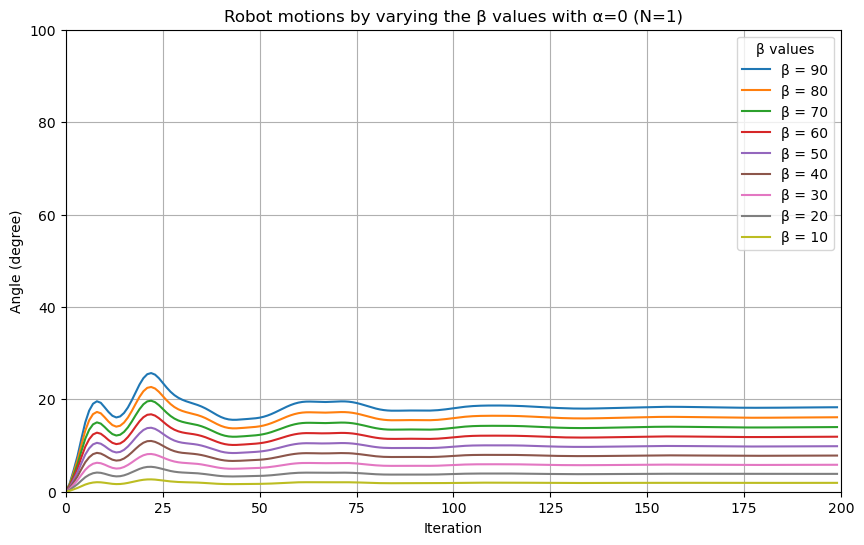}
        (c) $\beta$ over time
    \end{minipage}
\caption{Actual motions with N=1.}\label{fig:motion_n1}
    
    \vspace{2em}
    
    \begin{minipage}[c]{0.32\textwidth}
        \centering
        \includegraphics[width=\textwidth]{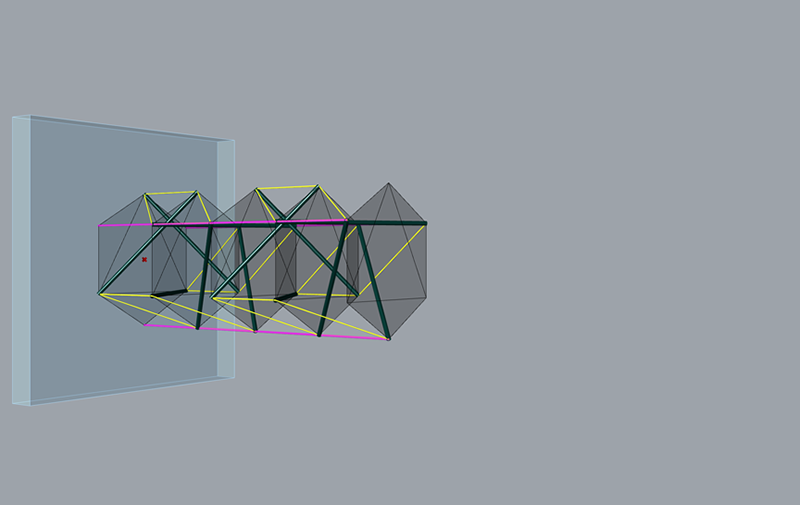}
        (a) N=2
    \end{minipage}
    \begin{minipage}[c]{0.32\textwidth}
        \centering
        \includegraphics[width=\textwidth]{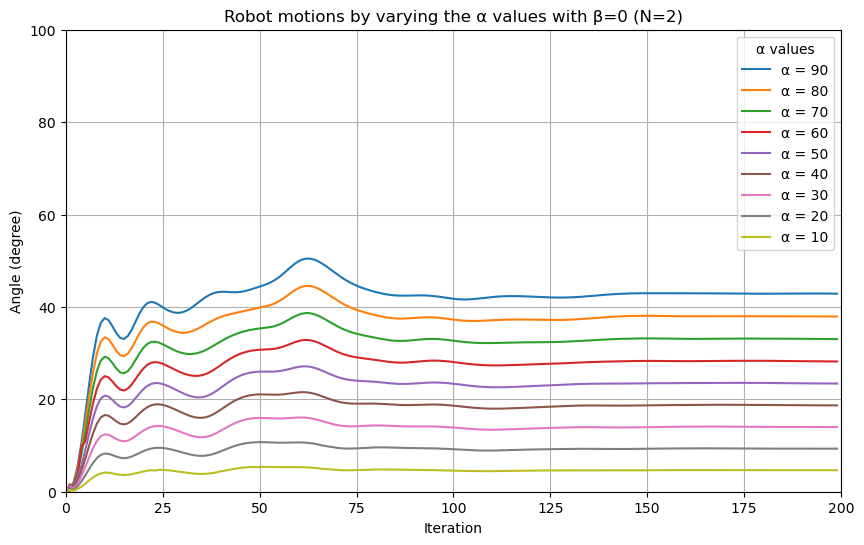}
        (b) $\alpha$ over time
    \end{minipage}
    \begin{minipage}[c]{0.32\textwidth}
        \centering
        \includegraphics[width=\textwidth]{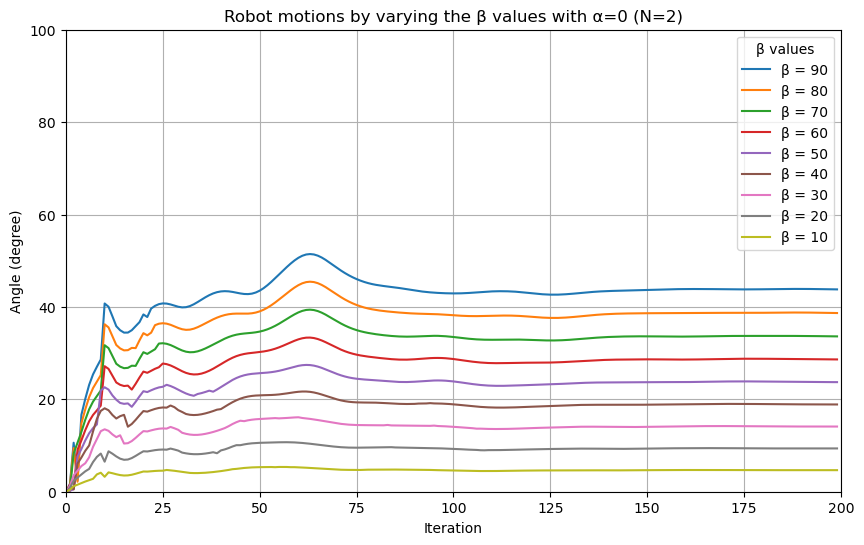}
        (c) $\beta$ over time
    \end{minipage}
\caption{Actual motions with N=2.}\label{fig:motion_n2}

    \vspace{2em}
    
    \begin{minipage}[c]{0.32\textwidth}
        \centering
        \includegraphics[width=\textwidth]{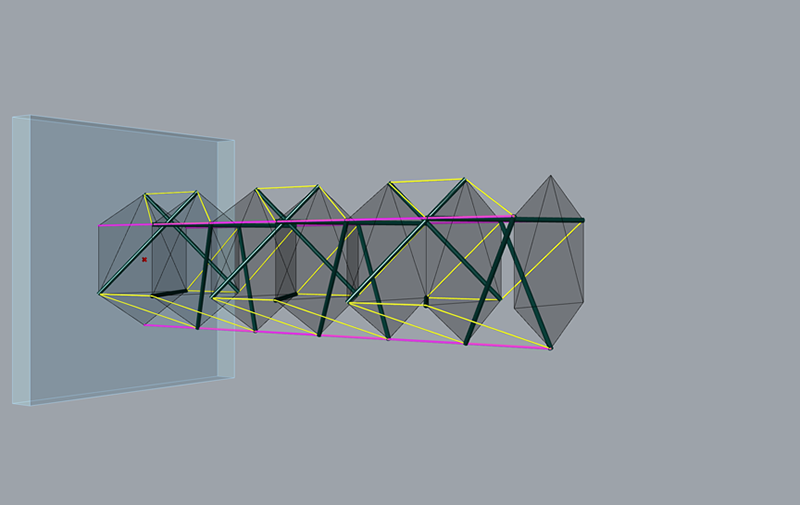}
        (a) N=3.
    \end{minipage}
    \begin{minipage}[c]{0.32\textwidth}
        \centering
        \includegraphics[width=\textwidth]{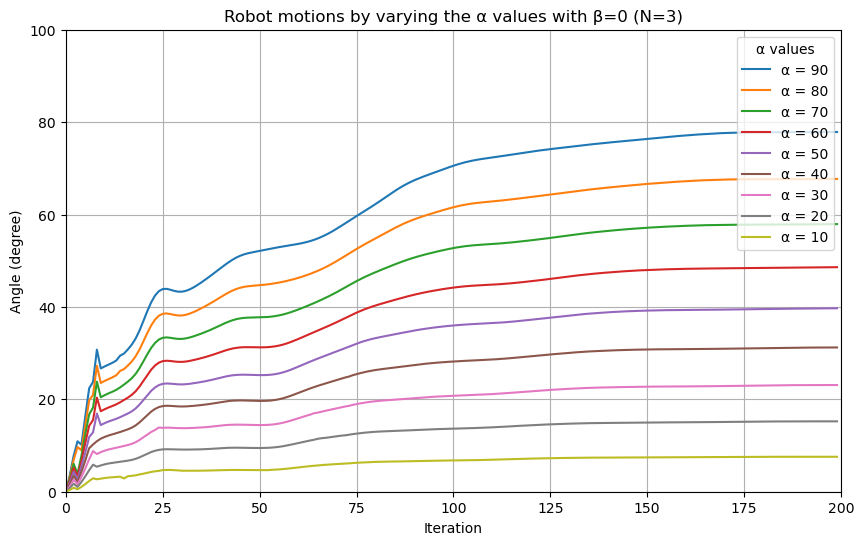}
        (b) $\alpha$ over time
    \end{minipage}
    \begin{minipage}[c]{0.32\textwidth}
        \centering
        \includegraphics[width=\textwidth]{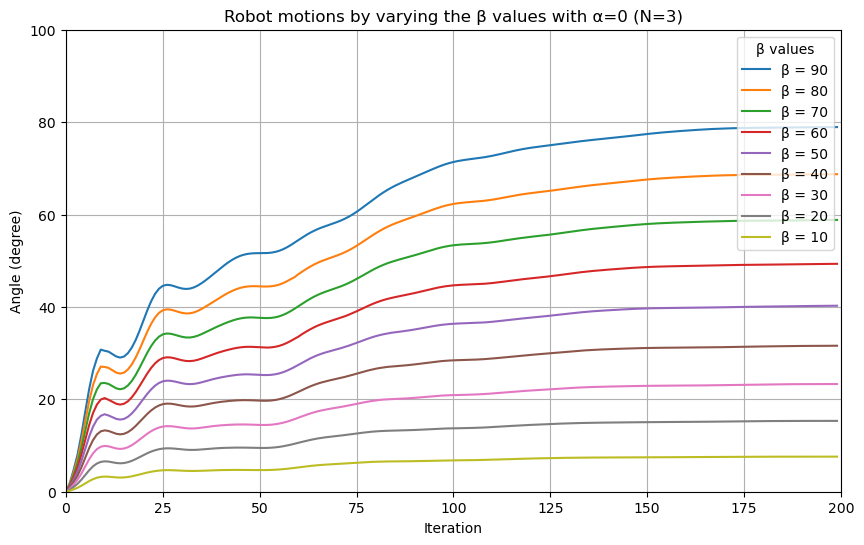}
        (c) $\beta$ over time
    \end{minipage}
\caption{Actual motions with N=3.}\label{fig:motion_n3}

    \vspace{2em}
    
    \begin{minipage}[c]{0.32\textwidth}
        \centering
        \includegraphics[width=\textwidth]{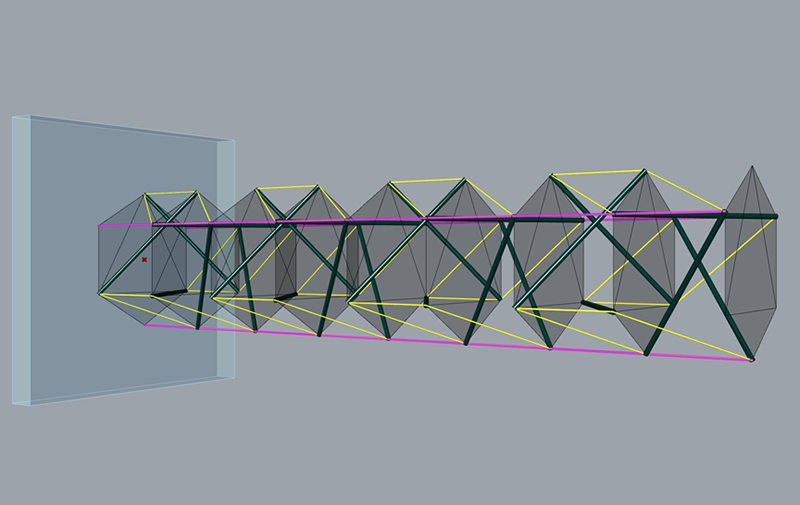}
        (a) N = 4
    \end{minipage}
    \begin{minipage}[c]{0.32\textwidth}
        \centering
        \includegraphics[width=\textwidth]{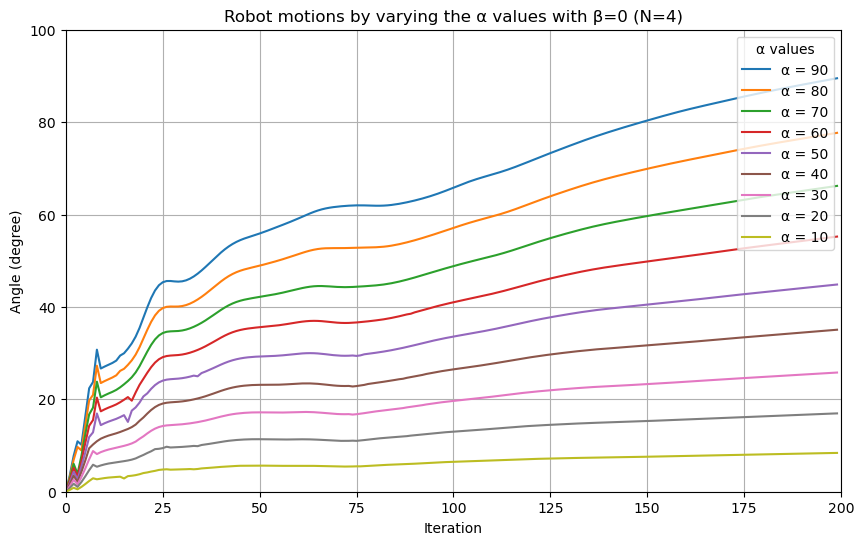}
        (b) $\alpha$ over time
    \end{minipage}
    \begin{minipage}[c]{0.32\textwidth}
        \centering
        \includegraphics[width=\textwidth]{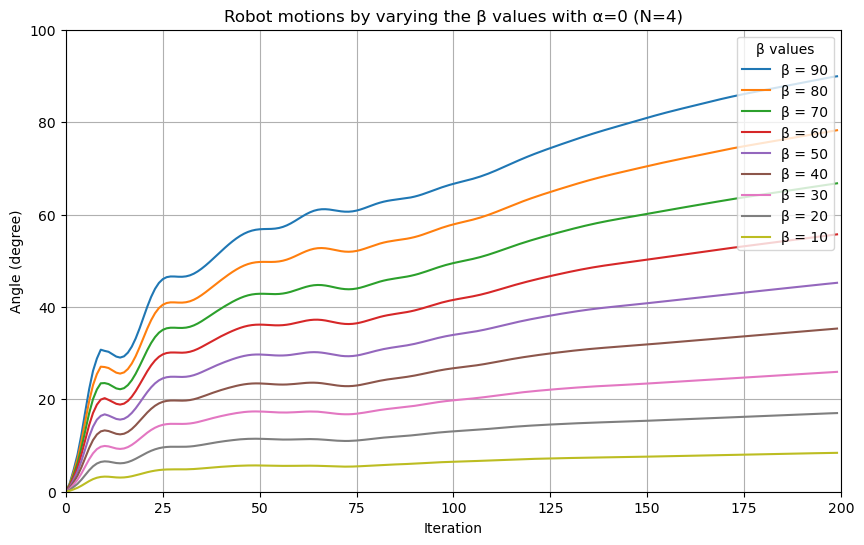}
        (c) $\beta$ over time
    \end{minipage}

\caption{Actual motions with N=4.}\label{fig:motion_n4}
    
\end{figure}

To achieve a better control, we utilize the transfer function proposed by \citet{relano2022modeling}. This transfer function focuses on tendon-driven continuum robots with three tendon actuators, which is applicable in this context. It is demonstrated by a decoupling mathematical process, aiming to identify the equations of control inputs and action outputs. The variables in this transfer function are yaw and pitch angles ($\alpha$, $\beta$) and corresponding variations of the tendon lengths ($L_1$, $L_2$, $L_3$), assuming two degrees of freedom (DOF) regarding motion capabilities. Additionally, it is also assumed that $L_1$+$L_2$+$L_3$=0, indicating the entire continuum robot remains the same angle of curvature while bending. Given these assumptions, the transfer function can be described as:
\par
\begin{align}
    L_1 &= \frac{\alpha}{1.5} \\
    L_2 &= \frac{\beta}{1.732} - \frac{\alpha}{3} \\
    L_3 &= -\frac{\beta}{1.732} - \frac{\alpha}{3}
\end{align}

\par To validate the actual performance, we have applied these control algorithms to the robot simulation. The validation experiment is conducted by inserting a series of target yaw and pitch angles from -90$^{\circ}$ to 90$^{\circ}$ at an interval of 10, and collecting the length variation values of each tendon actuator. The evaluation of yaw and pitch movements are divided into two seperate processes, which means that the simulation inputs $\alpha$ and $\beta$ are 
non-zero and zero value alternatively. This allows the vertical and horizontal movements to be clearly viewed seperately. Four conditions with different number of segments (N=1, 2, 3, 4) are selected as the subjects to test out the impact of value N on the actual movements. Figure~\ref{fig:motion_n1}, \ref{fig:motion_n2}, \ref{fig:motion_n3}, and \ref{fig:motion_n4} illustrate the robot's actual motions over time (t), with (a) the physical configuration with t=0, (b) the variation of yaw angle ($\alpha$), and (c) the pitch angle ($\beta$). The performance analysis reveals several key observations:

\begin{itemize}
    \item \textbf{Impact of segment number}: Increasing the number of segments (N) generally improves the robot's ability to achieve the desired motions. This is attributed to the increased degrees of freedom provided by additional segments, allowing for more precise control of the yaw and pitch angles.Configurations with N=1 to N=3 inherently exhibit limited motion ranges due to the restricted degrees of freedom. These configurations struggle to meet the desired yaw and pitch angles accurately.
    \item \textbf{Trade-offs with increased segments}: While adding more segments enhances performance, it also introduces greater structural complexity. The influence of gravitational loads become more pronounced as well, potentially leading to stability issues and an increased computational load.
    \item \textbf{Optimal performance with N=4}: The configuration with four segments demonstrates the best performance, closely approximating the desired motions. This configuration strikes a balance between flexibility and structural complexity, making it the most efficient setup for the following research.
\end{itemize}

\begin{figure}[htb]
    \centering
    \begin{minipage}[t]{.48\linewidth}
        \centering
        \includegraphics[width=\linewidth]{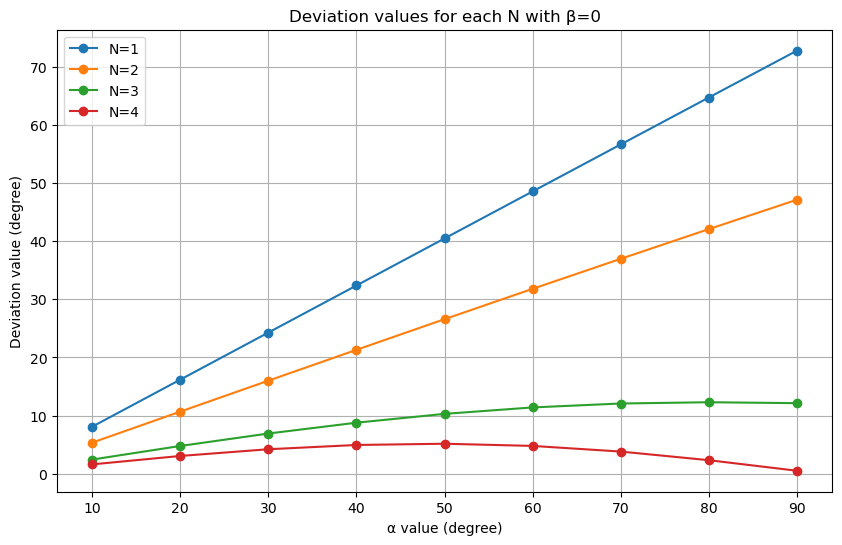}
        (a) $\alpha$ deviations (with $\beta$=0)
    \end{minipage}\hfill
    \begin{minipage}[t]{.48\linewidth}
        \centering
        \includegraphics[width=\linewidth]{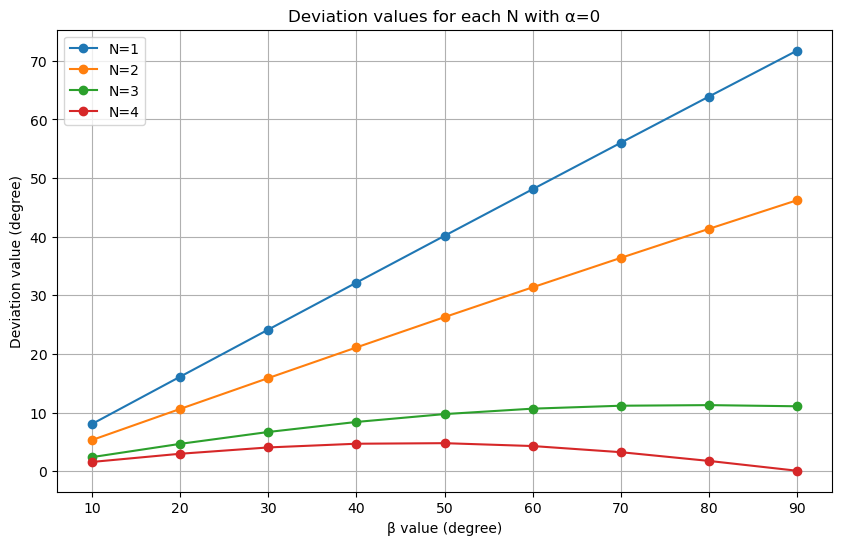}
        (b) $\beta$ deviations (with $\alpha$=0)
    \end{minipage}
\caption{Deviations for each N.}\label{fig:main_caption_deviation}
\end{figure}

\par However, even with the optimal configuration (N=4), there are still observable deviations between the actual and desired motions (Figure \ref{fig:main_caption_deviation}). These discrepancies indicate that the original transfer functions are not fully adequate for capturing the non-linear dynamics of the bionic tendon-driven robot. In fact, it requires a better optimization approach to adjust the coefficients in the function. To address this issue, the following sections explore the use of machine learning models to derive more accurate and robust transfer functions. By leveraging data-driven approaches, we aim to improve the control and modeling of the robot's dynamics, thereby enhancing its overall performance.    

\section{Methods}

\begin{figure}[htbp]
    \begin{center}
        \includegraphics[width=\linewidth]{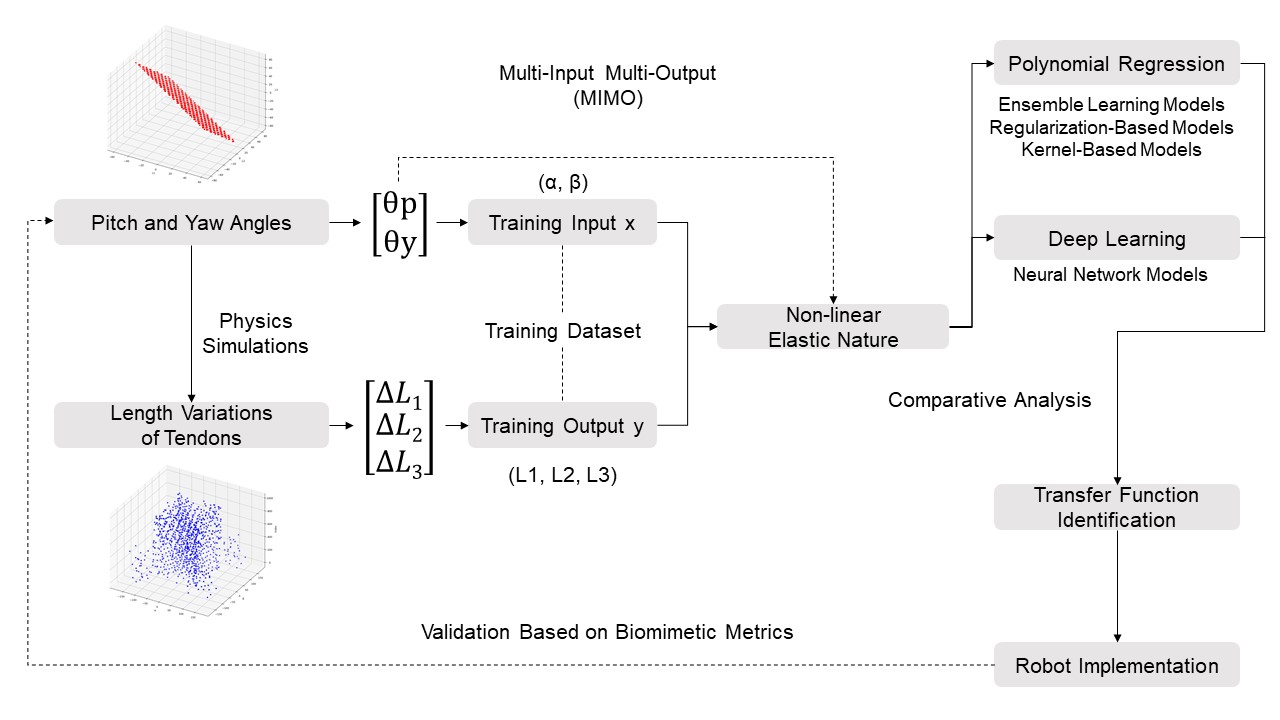}
        \caption{Methodology.}\label{fig:fig_methodology}
    \end{center}
\end{figure}

The proposed research methodology is comprised of four principal phases (Figure \ref{fig:fig_methodology}). In the preliminary phase, data collection is conducted using physics simulations to capture robotic motions (yaw and pitch angles) and corresponding actuation parameters (length variations of each tendon). Subsequently, a variety of learning models suitable for non-linear data are selected and trained with these datasets. A comparative analysis is then conducted to evaluate the performance of each model, including accuracy and computation costs. Eventually, the transfer functions derived from each learning model are sent back to the robot simulation to validate the actual locomotion based on biomimetic metrics.

\subsection{Data collection}
\begin{figure}[htb]
    \centering
    \begin{minipage}[t]{.5\linewidth}
        \centering
        \includegraphics[width=\linewidth]{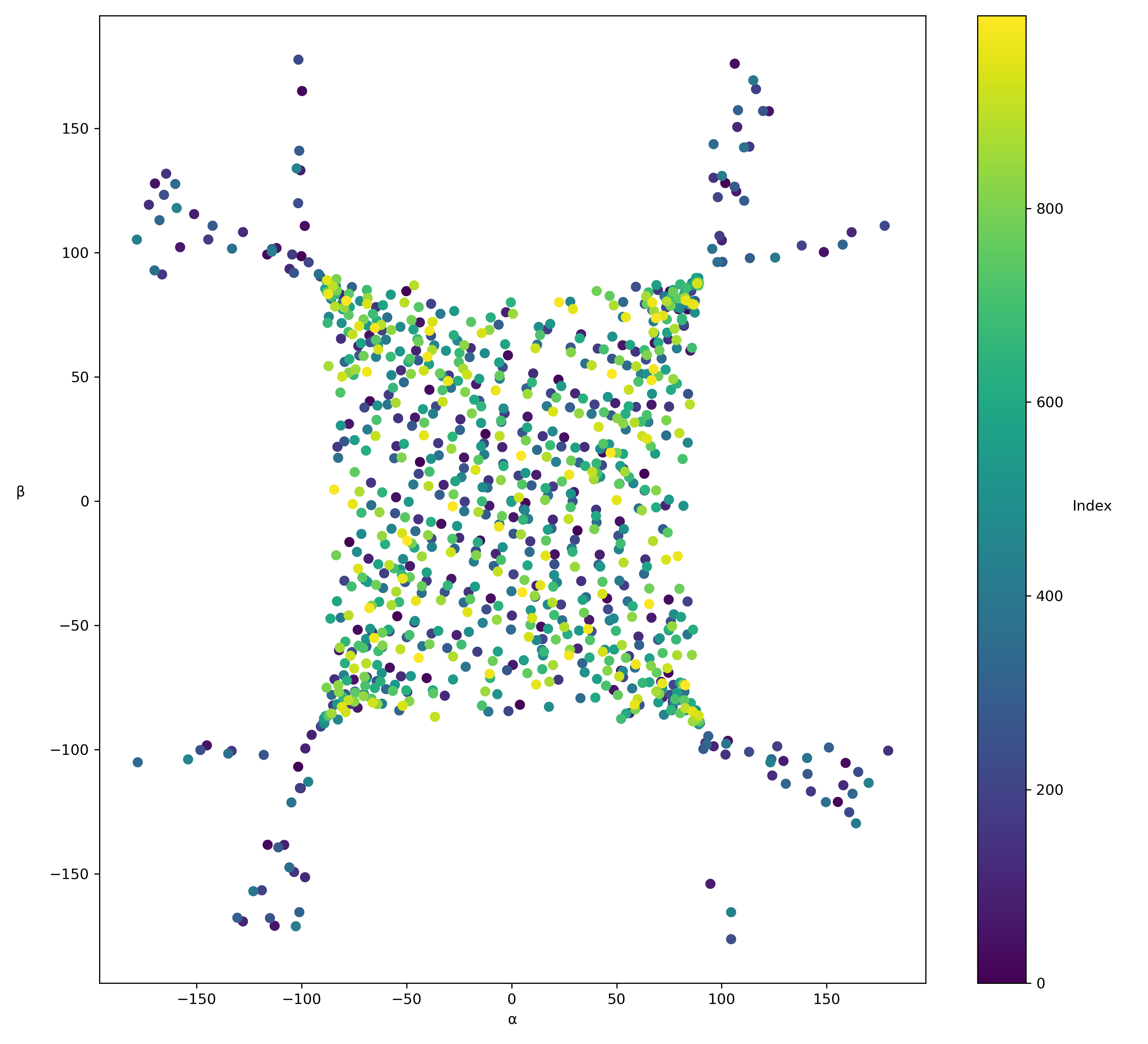}
        (a) Inputs: Pitch and yaw angles
    \end{minipage}\hfill
    \begin{minipage}[t]{.5\linewidth}
        \centering
        \includegraphics[width=\linewidth]{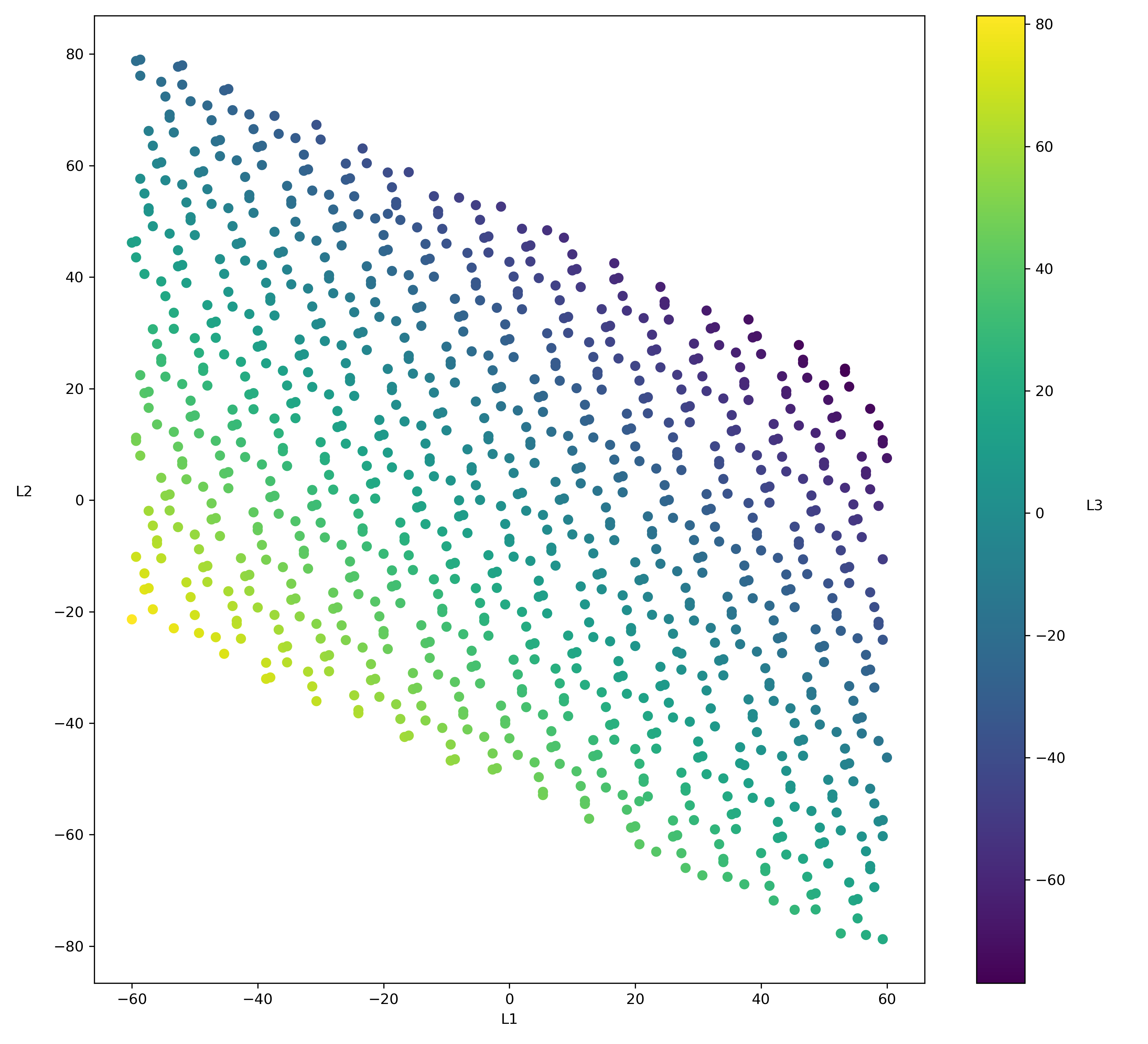}
        (b) Outputs: Length variations of tendons
    \end{minipage}
\caption{Training data collection.}\label{fig:main_caption_dataCollection}
\end{figure}

The training dataset for this research was constructed to capture the complex dynamics (in full range of possible motions) of the robot in a multi-input multi-output (MIMO) condition. The data collection process (Figure \ref{fig:main_caption_dataCollection}) involved simulating various pitch (\(\alpha\)) and yaw (\(\beta\)) angles and recording the corresponding length factors ($L_1$, $L_2$, $L_3$).

\par To ensure the robustness and generalization of the training dataset, the input angles were varied systematically, and multiple measurements were taken for each angle combination. This approach ensures that the machine learning models trained on this dataset can generalize well to unseen data, thereby enhancing their predictive performance and reliability.

\par This process was conducted using a physics-based computational environment Grasshopper, which is a computer-aided design (CAD) platform. It allows for simulations of the robot's movements with intuitive control and precise measurement, ensuring high-fidelity data and reliable validations.

\subsection{Learning model selection}
In this section, we evaluate the performance of eight different machine learning models to derive transfer functions for bionic tendon-driven robots. The selected models are categorized into four sets: ensemble learning models, regularization-based models, kernel-based models, and neural network models. These models were selected based on their suitability for handling non-linear, MIMO data, which can be summarized as the following:

\subsection*{Ensemble Learning Models}

\subsubsection{Random Forest (RF)}
\textbf{Mathematical basis:} Random Forest is an ensemble learning method that builds multiple decision trees and merges their outcomes to improve accuracy and prevent overfitting. It is robust to noise and can capture non-linear relationships. \\

\subsubsection{Gradient Boosting (GB)}
\textbf{Mathematical basis:} Gradient Boosting builds an ensemble of weak prediction models, typically decision trees, and optimizes them iteratively to minimize the loss function. It is effective for capturing intricate patterns in data and handling non-linear dependencies. \\

\subsection*{Regularization-Based Models}
\subsubsection{Lasso Regression}
\textbf{Mathematical basis:} Lasso Regression (Least Absolute Shrinkage and Selection Operator) is a linear model that performs both variable selection and regularization, enhancing prediction accuracy and interpretability. It is useful for high-dimensional datasets where some features may be irrelevant. \\

\subsubsection{Ridge Regression}
\textbf{Mathematical basis:} Ridge Regression addresses multicollinearity issues by adding an \( l_2 \) penalty to the loss function. It shrinks the model coefficients, making it more robust to overfitting and capable of handling correlated features. \\

\subsection*{Kernel-Based Models}

\subsubsection{Support Vector Regressor (SVR)}
\textbf{Mathematical basis:} SVR extends support vector machines to regression problems. It is effective for high-dimensional spaces and can model non-linear relationships through kernel functions. \\

\subsubsection{Gaussian Process Regressor (GPR)}
\textbf{Mathematical basis:} GPR provides a probabilistic approach to regression, offering not only predictions but also uncertainty estimates. It is highly flexible and can model complex non-linear relationships. \\

\subsection*{Neural Network Models}

\subsubsection{Bayesian Neural Network (BNN)}
\textbf{Mathematical basis:} BNN incorporates Bayesian inference into neural networks, providing probabilistic predictions and uncertainty estimates. It is robust to overfitting and can model non-linear relationships effectively. \\

\subsubsection{Recurrent Neural Network (RNN)}
\textbf{Mathematical basis:} RNNs are designed to handle sequential data and temporal dependencies, making them suitable for modeling dynamic systems. They can capture the time-dependent behavior of tendon-driven robots. \\

\section{Results}

\subsection{Performance evaluation}

\begin{figure}[htp]
    \centering
    \begin{minipage}[t]{.24\textwidth}
        \centering
        \includegraphics[width=.8\textwidth]{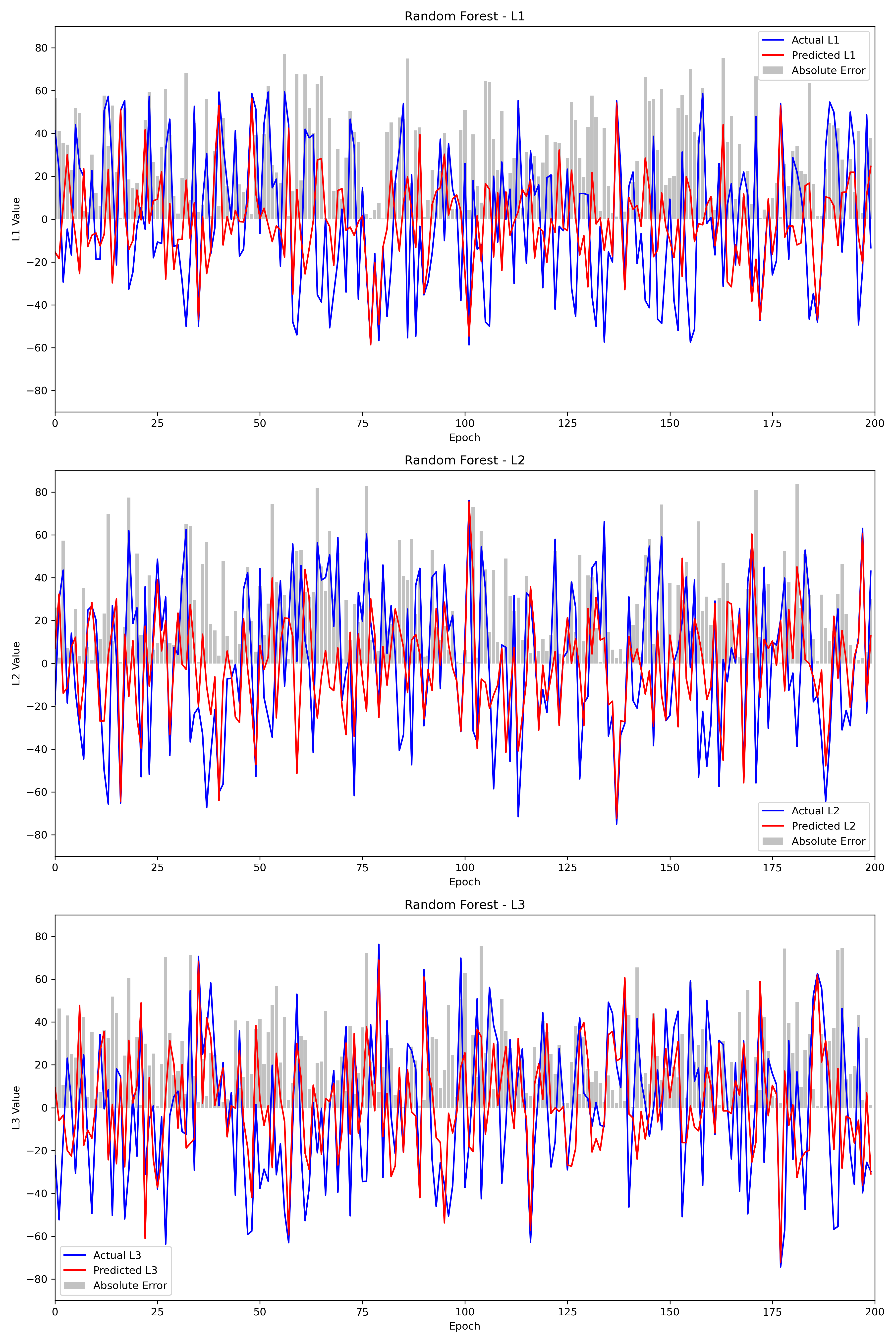}
        \caption{RF.}\label{fig:fig_Random_Forest_results}
    \end{minipage}
    \hfill
    \begin{minipage}[t]{.24\textwidth}
        \centering
        \includegraphics[width=.8\textwidth]{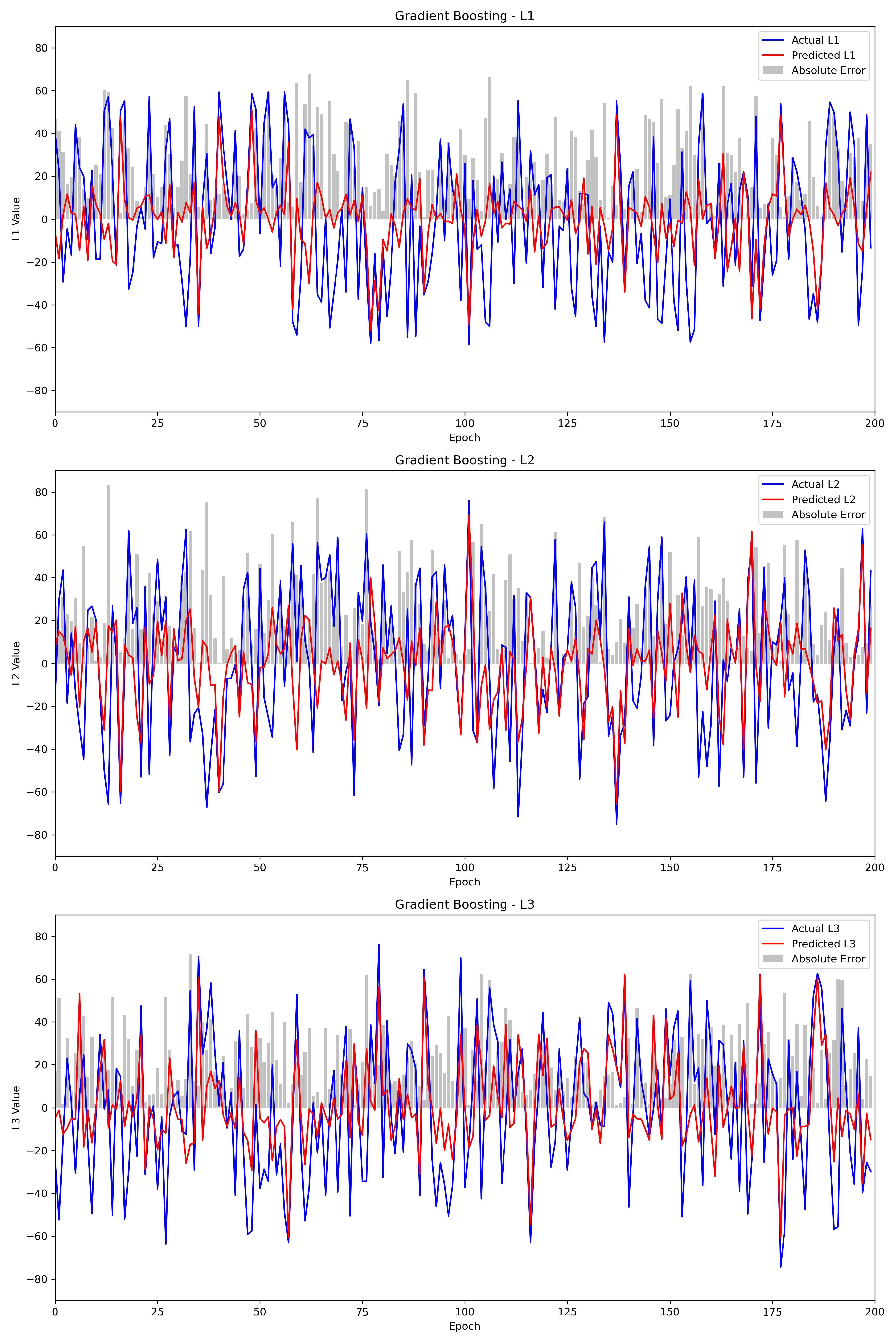}
        \caption{GB.}\label{fig:fig_Gradient_Boosting_results}
    \end{minipage}
    \hfill
    \begin{minipage}[t]{.24\textwidth}
        \centering
        \includegraphics[width=.8\textwidth]{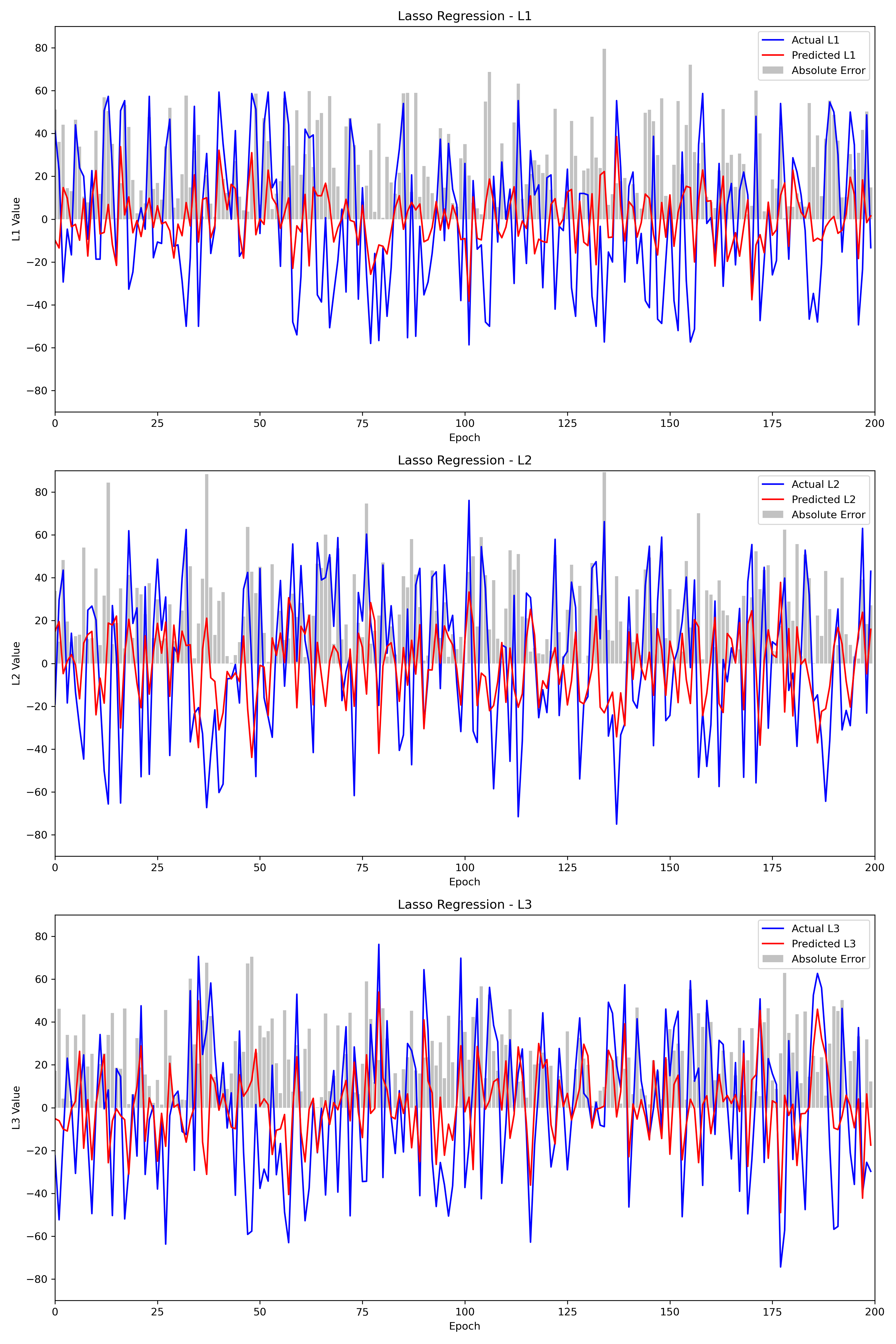}
        \caption{Lasso.}\label{fig:fig_Lasso_Regression_results}
    \end{minipage}
    \hfill
    \begin{minipage}[t]{.24\textwidth}
        \centering
        \includegraphics[width=.8\textwidth]{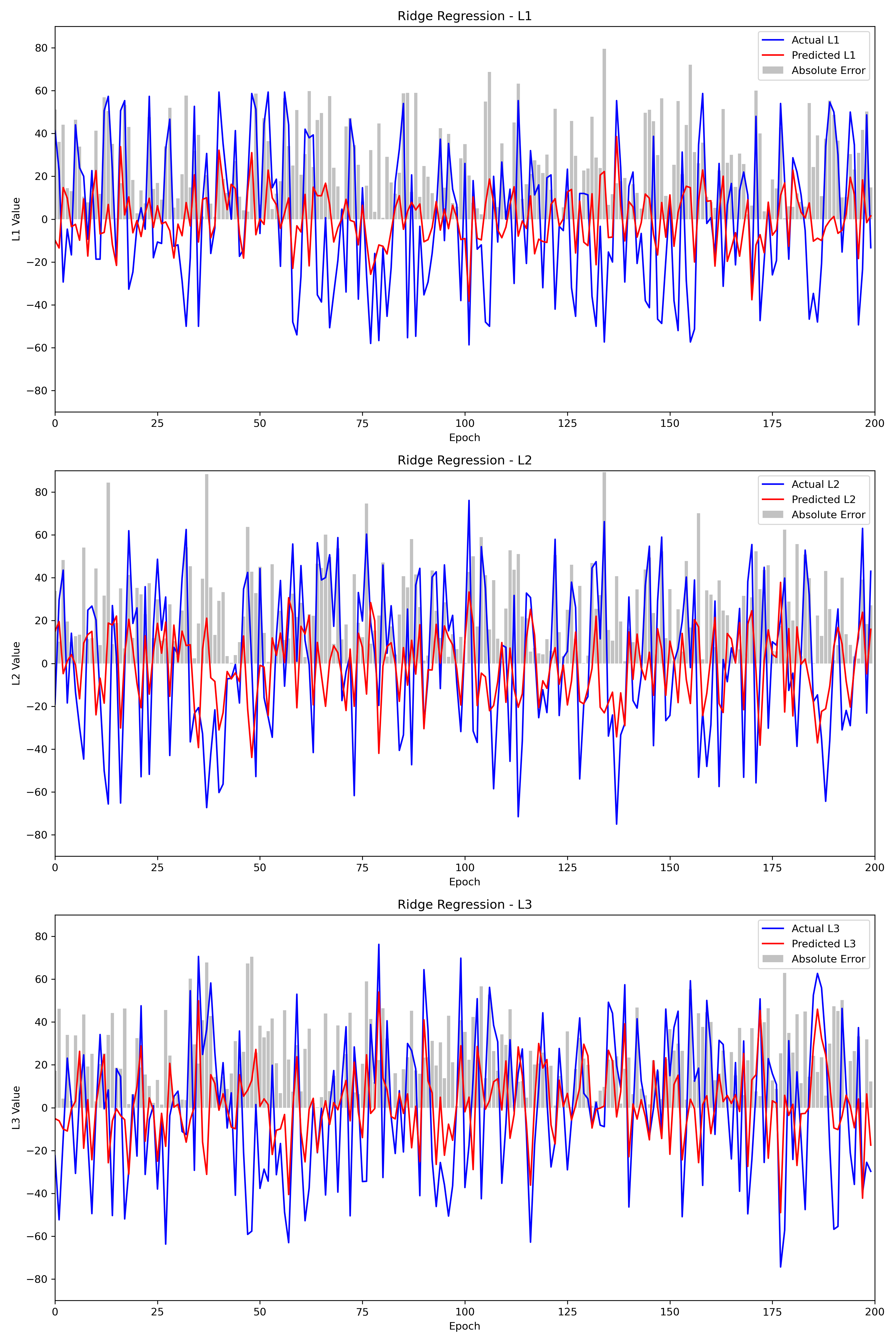}
        \caption{Ridge.}\label{fig:fig_Ridge_Regression_results}
    \end{minipage}
\end{figure}

\begin{figure}[htp]
    \centering
    \begin{minipage}[t]{.24\textwidth}
        \centering
        \includegraphics[width=.85\textwidth]{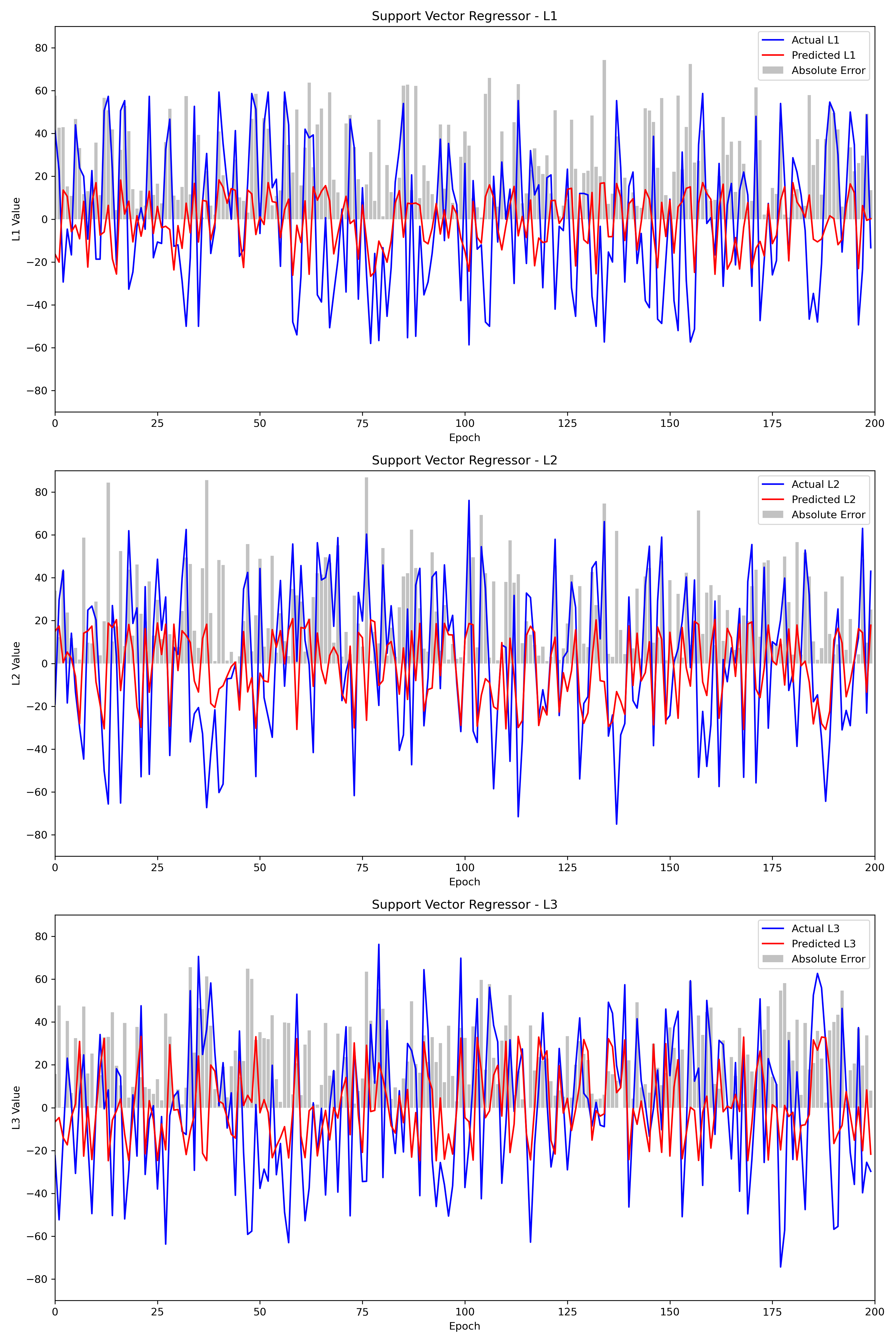}
        \caption{SVM.}\label{fig:fig_Support_Vector_Regressor_results}
    \end{minipage}
    \hfill
    \begin{minipage}[t]{.24\textwidth}
        \centering
        \includegraphics[width=.85\textwidth]{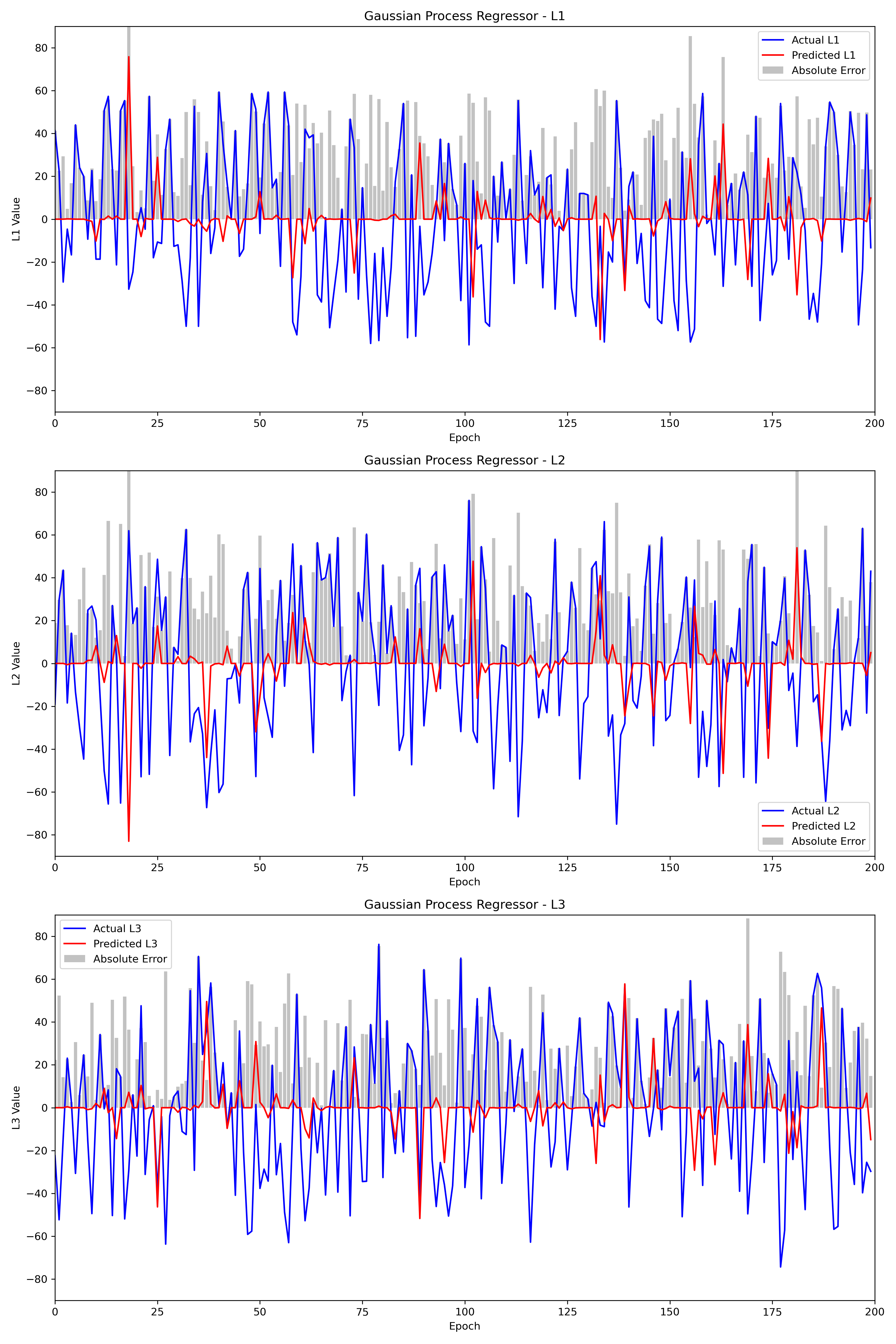}
        \caption{GPR.}\label{fig:fig_Gaussian_Process_Regressor_results}
    \end{minipage}
    \hfill
    \begin{minipage}[t]{.24\textwidth}
        \centering
        \includegraphics[width=.85\textwidth]{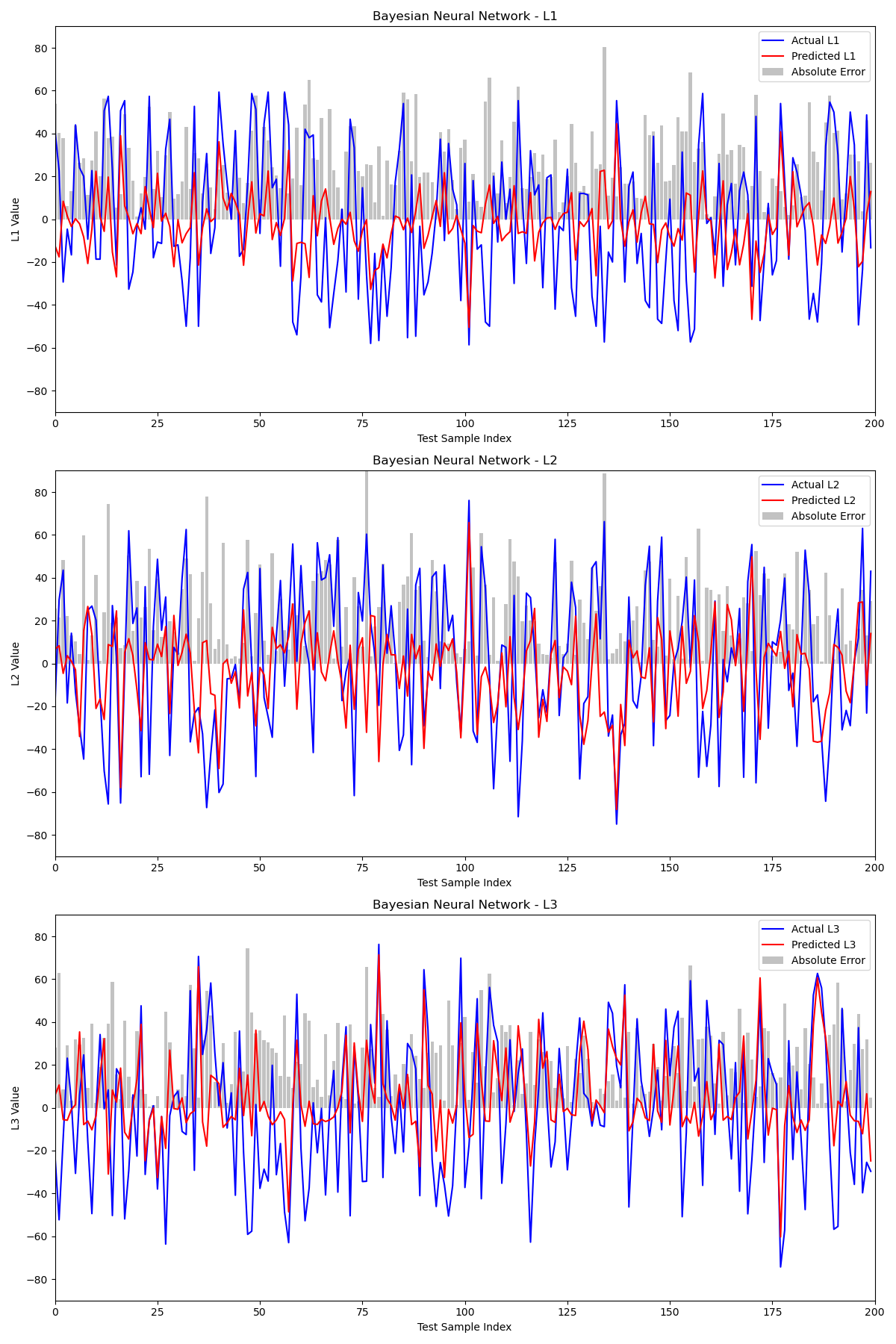}
        \caption{BNN.}\label{fig:fig_Bayesian_Neural_Network_results}
    \end{minipage}
    \hfill
    \begin{minipage}[t]{.24\textwidth}
        \centering
        \includegraphics[width=.85\textwidth]{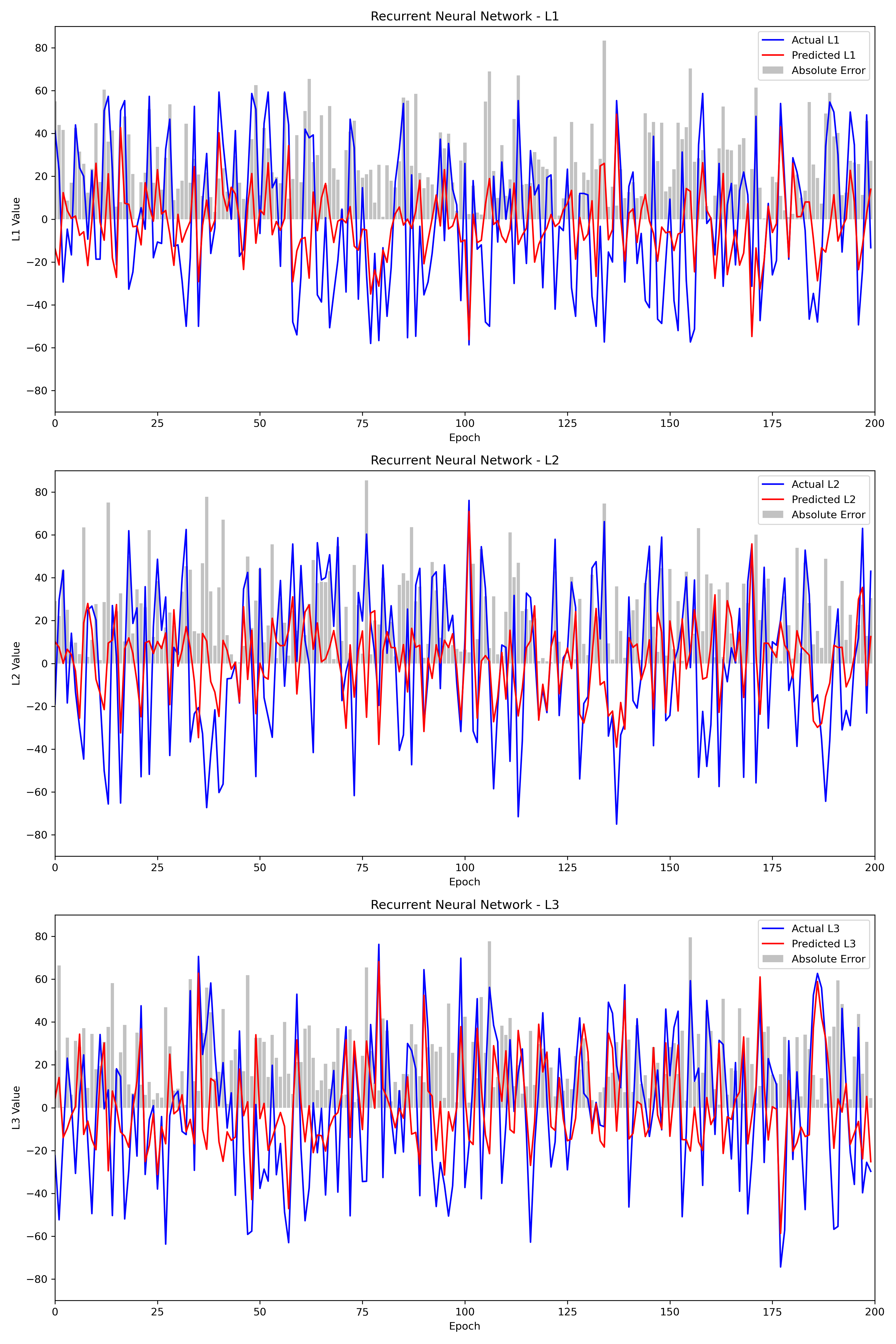}
        \caption{RNN.}\label{fig:fig_Recurrent_Neural_Network_results}
    \end{minipage}
\end{figure}

The training process (Figure~\ref{fig:fig_Random_Forest_results} to~\ref{fig:fig_Recurrent_Neural_Network_results}) for each learning model is evaluated by comparing the predicted values (red) against the actual values (blue) for L1, L2, and L3, along with the absolute error (gray).
\par For RF (Figure \ref{fig:fig_Random_Forest_results}) and GPR models (Figure \ref{fig:fig_Gaussian_Process_Regressor_results}), the predictions for L1, L2, and L3 exhibit high variability with significant deviations between the predicted and actual values, indicating considerable errors across all samples. The predictions from GB model (Figure \ref{fig:fig_Gradient_Boosting_results}) are closely aligned with the actual values, indicating superior accuracy and stability. It demonstrate the best performance among the models. For regularization-based models, the predictions from Lasso (Figure \ref{fig:fig_Lasso_Regression_results}) and Ridge (Figure \ref{fig:fig_Ridge_Regression_results}) regression models both show variability and deviations. The absolute error indicates moderate performance, with some significant errors. For SVR model (Figure \ref{fig:fig_Support_Vector_Regressor_results}), the predictions show moderate alignment with the actual values, suggesting a balanced performance. The predictions from BNN (Figure \ref{fig:fig_Bayesian_Neural_Network_results}) and RNN (Figure \ref{fig:fig_Recurrent_Neural_Network_results}) are closely aligned with the actual values, demonstrating a more stable performance and high accuracy.

\renewcommand{\arraystretch}{1.2}
\begin{longtable}{lccc}
    \caption{Performance metrics of different learning models} \label{table_evaluation} \\
    \hline
    \textbf{Model} & \textbf{MSE} & \textbf{MAE} & \textbf{Computation Time (s)} \\
    \hline
    Random Forest & 1058.216120 & 25.694405 & 1.760020 \\
    Gradient Boosting & 866.197726 & \textbf{23.549753} & 0.526898 \\
    Ridge & 927.343365 & 24.714297 & 0.013705 \\
    Lasso & 927.339126 & 24.714440 & \textbf{0.013146} \\
    Support Vector & 935.759171 & 24.398921 & 0.243148 \\
    Gaussian Processes & 1243.854615 & 29.352238 & 1.734590 \\
    BNN & 882.995994 & 24.169767 & 28.626474 \\
    RNN & 876.559293 & \textbf{23.967265} & 32.470035 \\
    \hline
\end{longtable}

\vspace{1cm}

\par Table~\ref{table_evaluation} presents an overall performance of each learning model, including mean square error (MSE), mean absolute error (MAE), and computation time. While GB and SVR exhibit significant scores in MAE, their counterparts RF and GPR show worse performance, indicating the instability of ensemble learning and kernel-based models for bionic robots. Neural network models BNN and RNN, on the other hand, demonstrate a more stable solution with high accuracy. However, it requires more computational complexity. Regarding regularization-based models, despite their lower accuracy, Ridge and Lasso models exhibit significantly reduced computation time, making them efficient for real-time applications. This efficiency is crucial for bionic robots, where adaptability and responsiveness are more important than extreme precision, aligning with the goals of replicating biological movements.

\begin{figure}[H]
    \centering
    \begin{minipage}[t]{.49\textwidth}
        \centering
        \includegraphics[width=.9\textwidth]{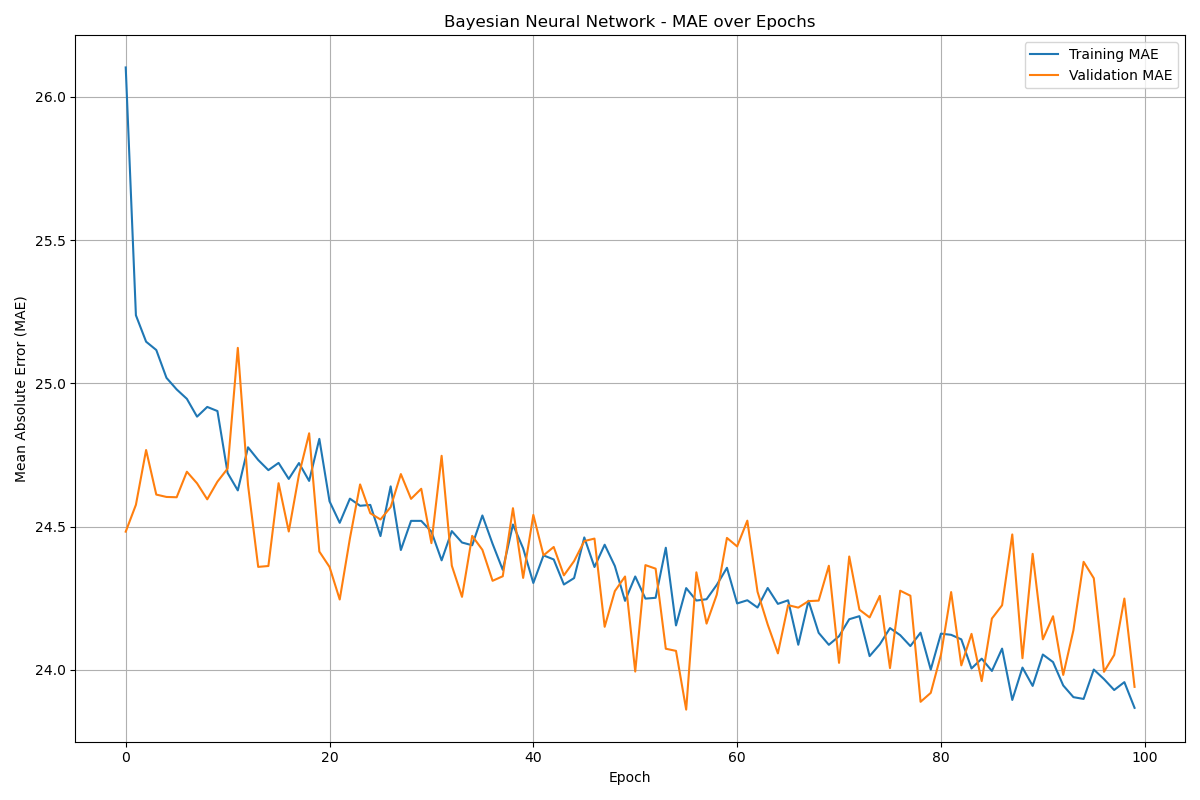}
        \caption{BNN: MAE over epochs.}\label{fig:fig_evaluation_Bayesian_Neural_Network_MAE_over_epochs}
    \end{minipage}
    \hfill
    \begin{minipage}[t]{.49\textwidth}
        \centering
        \includegraphics[width=.9\textwidth]{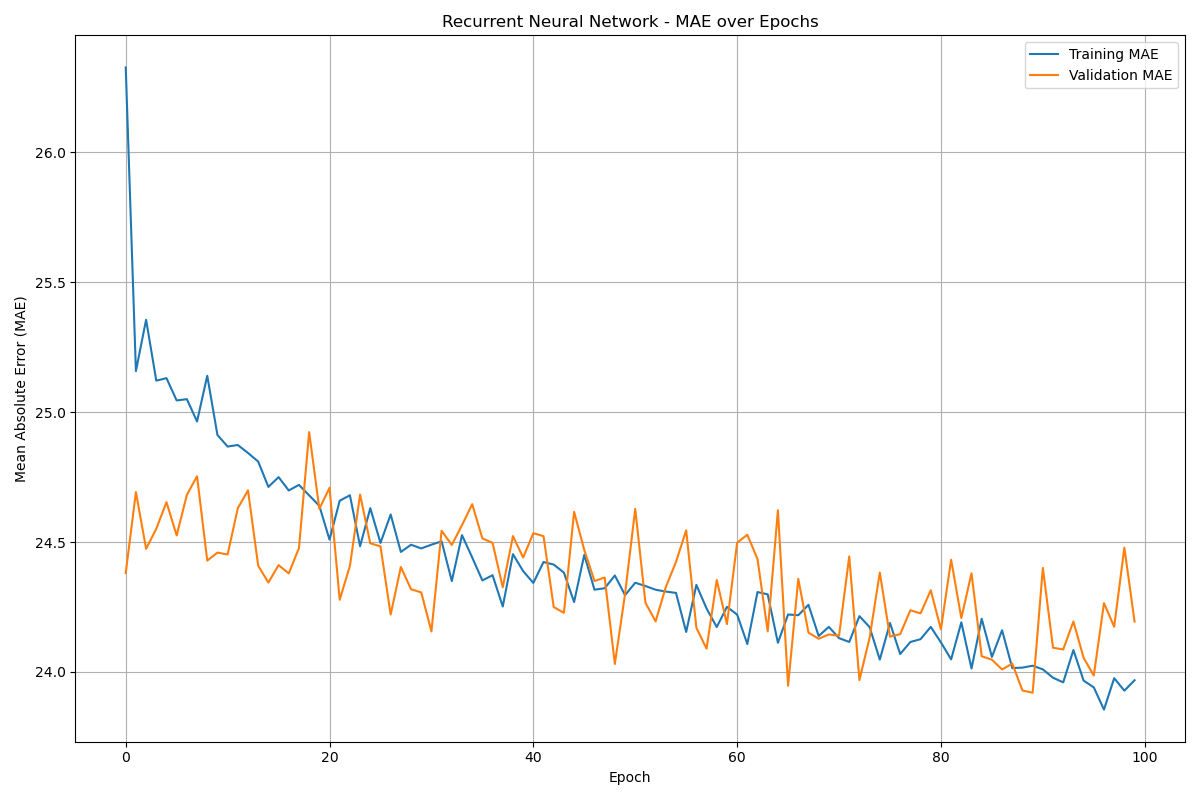}
        \caption{RNN: MAE over epochs.}\label{fig:fig_evaluation_Recurrent_Neural_Network_MAE_over_epochs}
    \end{minipage}
\end{figure}
    
\par The research further explore the generalization capabilities of the neural network models for robotic learning by presenting the training and validation MAE over 100 epochs (Figure~\ref{fig:fig_evaluation_Bayesian_Neural_Network_MAE_over_epochs} and~\ref{fig:fig_evaluation_Recurrent_Neural_Network_MAE_over_epochs}). Both training MAE of BNN and RNN exhibit a sharp decline during the initial 20 epochs, indicating that these models can quickly learn from the training data. The BNN model displays a stable validation MAE curve, while the RNN model shows slight fluctuations, reflecting its sensitivity to temporal variations in the data. Overall, both models are capable of capturing the complex dynamics of bionic tendon-driven robots, making them suitable for developing accurate, robust model-free control strategies.

\subsection{Transfer function identification}

To capture non-linear relationships between the input variables ($\alpha$, $\beta$) and the output variables ($L_1$, $L_2$, $L_3$), polynomial features are generated. For a second-degree polynomial, the feature set includes:

\[\{1, \alpha, \beta, \alpha^2, \alpha\beta, \beta^2\}\]

Mathematically, the polynomial features $\Phi(\mathbf{x})$ for input vector $\mathbf{x} = [\alpha, \beta]^T$ can be expressed as:

\[\Phi(\mathbf{x}) = [1, \alpha, \beta, \alpha^2, \alpha\beta, \beta^2]^T\]

Given the polynomial features, we fit a linear regression model to represent the non-linear relationship between the inputs and outputs. The regression model for each output $L_i$ (where $i \in \{1, 2, 3\}$) can be formulated as:

\[L_i = w_{i0} + w_{i1}\alpha + w_{i2}\beta + w_{i3}\alpha^2 + w_{i4}\alpha\beta + w_{i5}\beta^2\]

where $w_{ij}$ are the coefficients of the polynomial regression model for the $i$-th output and $j$-th term.

The coefficients obtained from the polynomial regression model are used to construct the transfer function equations. For each output $L_i$, the transfer function $T_i$ can be expressed as:

\[T_i(\alpha, \beta) = w_{i0} + w_{i1}\alpha + w_{i2}\beta + w_{i3}\alpha^2 + w_{i4}\alpha\beta + w_{i5}\beta^2\]

The process of identifying transfer functions using polynomial regression is applied to the predictions made by each machine learning model. This involves the following steps:

\begin{enumerate}
    \item \textbf{Training the machine learning model}: Each machine learning model is trained to predict the outputs $\hat{L_1}, \hat{L_2}, \hat{L_3}$ from the inputs $\alpha, \beta$.

    \item \textbf{Generating polynomial features}: Polynomial features are generated from the inputs $\alpha, \beta$.

    \item \textbf{Fitting polynomial regression}: A polynomial regression model is fitted using the polynomial features and the predicted outputs from the machine learning model.

    \item \textbf{Constructing transfer functions}: The coefficients and intercepts from the polynomial regression model are used to construct the transfer function equations.
\end{enumerate}

Thus, the transfer functions of each regression model can be derived as:

\subsection*{Random Forest}
\begin{align}
L1 &= 1.0197 + (0.1833) \alpha + (-0.0700) \beta + (-0.0001) \alpha^2 + (0.0002) \alpha \beta + (-0.0001) \beta^2 \\
L2 &= -0.4349 + (-0.0089) \alpha + (0.2548) \beta + (0.0002) \alpha^2 + (-0.0004) \alpha \beta + (-0.0003) \beta^2 \\
L3 &= -0.5848 + (-0.1744) \alpha + (-0.1848) \beta + (-0.0001) \alpha^2 + (0.0003) \alpha \beta + (0.0004) \beta^2
\end{align}

\subsection*{Gradient Boosting}
\begin{align}
L1 &= 1.0197 + (0.1833) \alpha + (-0.0700) \beta + (-0.0001) \alpha^2 + (0.0002) \alpha \beta + (-0.0001) \beta^2 \\
L2 &= -0.4349 + (-0.0089) \alpha + (0.2548) \beta + (0.0002) \alpha^2 + (-0.0004) \alpha \beta + (-0.0003) \beta^2 \\
L3 &= -0.5848 + (-0.1744) \alpha + (-0.1848) \beta + (-0.0001) \alpha^2 + (0.0003) \alpha \beta + (0.0004) \beta^2
\end{align}

\subsection*{Ridge Regression}
\begin{align}
L1 &= 0.1783 + (0.1839) \alpha + (-0.0704) \beta \\
L2 &= -0.7694 + (-0.0106) \alpha + (0.2540) \beta \\
L3 &= 0.5911 + (-0.1733) \alpha + (-0.1835) \beta
\end{align}

\subsection*{Lasso Regression}
\begin{align}
L1 &= 0.1782 + (0.1839) \alpha + (-0.0704) \beta \\
L2 &= -0.7693 + (-0.0106) \alpha + (0.2540) \beta \\
L3 &= 0.5911 + (-0.1733) \alpha + (-0.1835) \beta
\end{align}

\subsection*{Support Vector Regressor}
\begin{align}
L1 &= 1.0197 + (0.1833) \alpha + (-0.0700) \beta + (-0.0001) \alpha^2 + (0.0002) \alpha \beta + (-0.0001) \beta^2 \\
L2 &= -0.4349 + (-0.0089) \alpha + (0.2548) \beta + (0.0002) \alpha^2 + (-0.0004) \alpha \beta + (-0.0003) \beta^2 \\
L3 &= -0.5848 + (-0.1744) \alpha + (-0.1848) \beta + (-0.0001) \alpha^2 + (0.0003) \alpha \beta + (0.0004) \beta^2
\end{align}

\subsection*{Gaussian Process Regressor}
\begin{align}
L1 &= 1.0197 + (0.1833) \alpha + (-0.0700) \beta + (-0.0001) \alpha^2 + (0.0002) \alpha \beta + (-0.0001) \beta^2 \\
L2 &= -0.4349 + (-0.0089) \alpha + (0.2548) \beta + (0.0002) \alpha^2 + (-0.0004) \alpha \beta + (-0.0003) \beta^2 \\
L3 &= -0.5848 + (-0.1744) \alpha + (-0.1848) \beta + (-0.0001) \alpha^2 + (0.0003) \alpha \beta + (0.0004) \beta^2
\end{align}

\begin{figure}[H]
\begin{center}
    \includegraphics[width=0.75\textwidth]{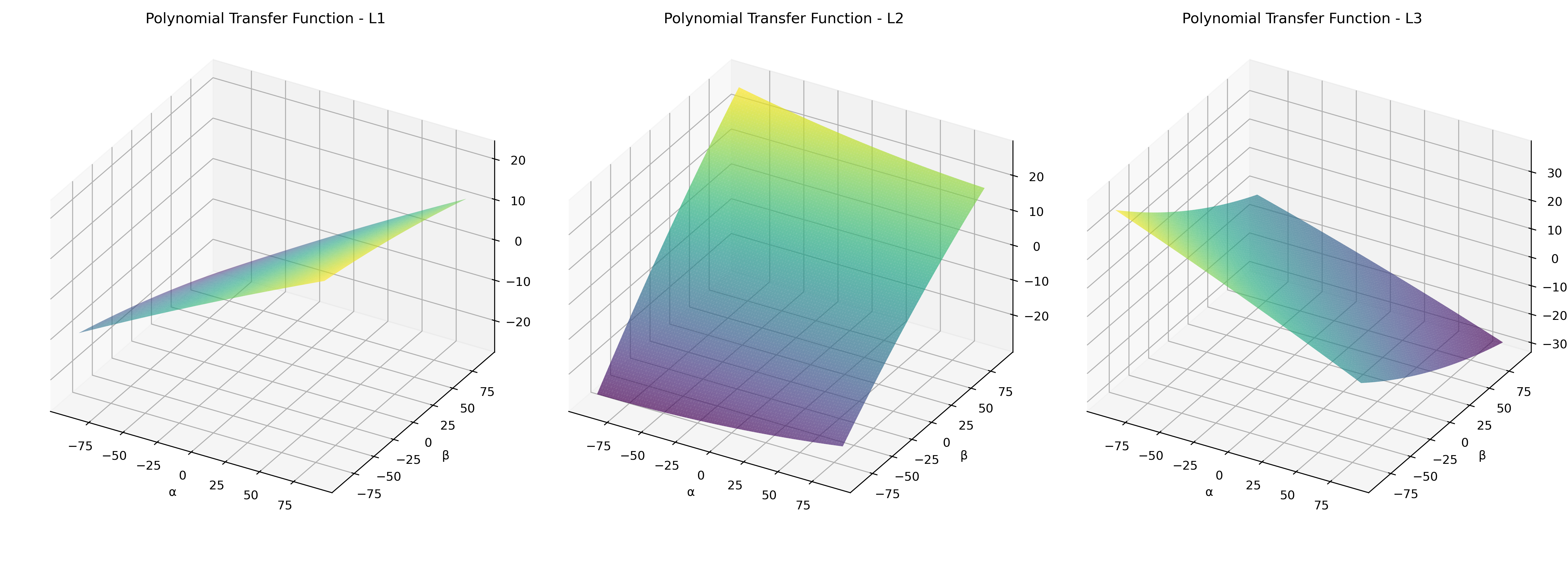}
    \caption{Optimal transfer functions.}\label{fig:fig_Polynomial_Transfer_Functions_3D}
\end{center}
\end{figure}

\begin{figure}[H]
\begin{center}
    \includegraphics[width=0.75\textwidth]{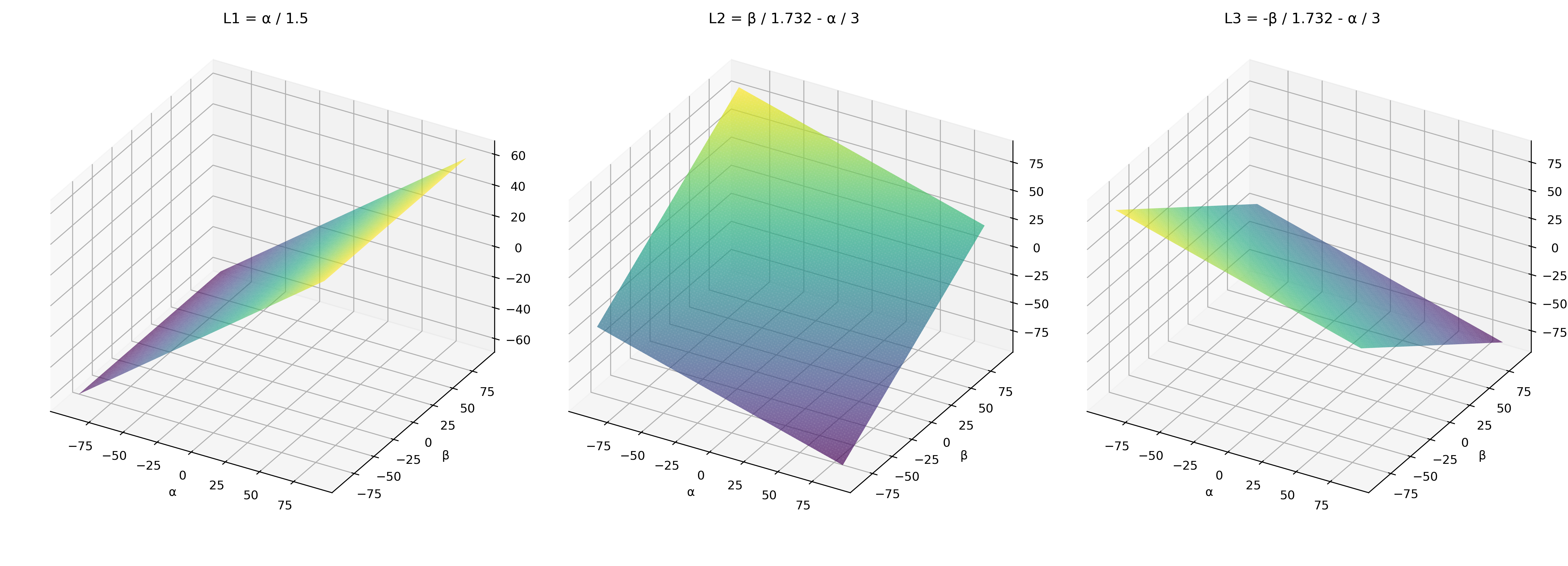}
    \caption{Original transfer functions.}\label{fig:fig_Original_Transfer_Functions_3D}
\end{center}
\end{figure}

Additionally, a comparative graph (Figure~\ref{fig:fig_Polynomial_Transfer_Functions_3D} and~\ref{fig:fig_Original_Transfer_Functions_3D}) is presented
to identify the difference between the optimial transfer functions (derived from GB model) and the original ones.
The changing in geometry can be considered as a fine-tuning result after the training process.

As for neural network models, BNN and RNN do not provide direct transfer functions due to their implicit nature (multiple layers of 
interconnected neurons and non-linear activation functions). Yet, they are still capable of predicting accurate control parameters.

\begin{figure}[htbp]
    \centering
    \includegraphics[width=.9\linewidth]{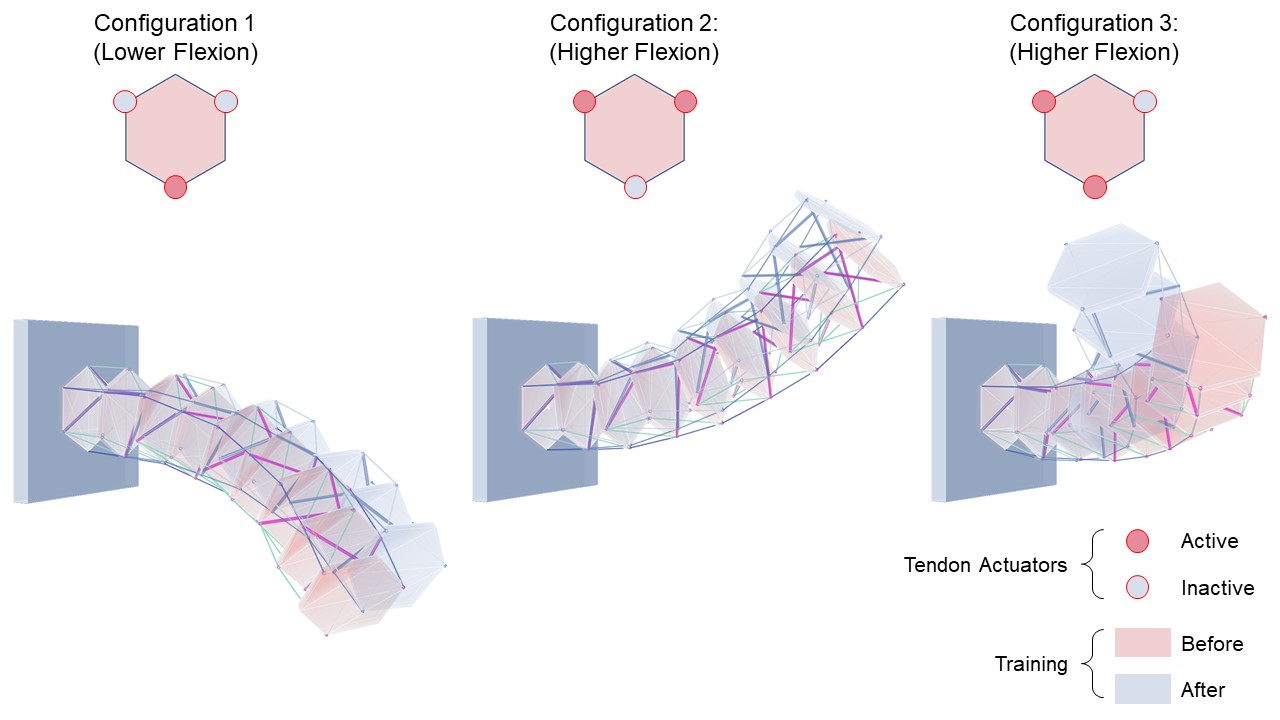} 
    \caption{Robot performance before and after training.}
    \label{fig:mlResult}
\end{figure}

\par The GB model, with the best performance among all learning models, was then fed back to the control system for the purpose of validating its actual impact on robot movements. Figure \ref{fig:mlResult} presents the configuration driven by the original transfer function (pink) and by the enhanced ones (blue). The results indicates that the robot can reduce the interference of gravitational loads and thus achieving more precise configurations.

\section{Conclusions}
This research focuses on the model-free methods for bionic robots using learning models to optimize the modeling and control of their complex dynamics. Due to the lack of a comprehensive evaluation of various machine learning methods for bionic robots, roboticists often encounter difficulties in identifying an effective learning model to optimize their specific robots. To address these challenges, a comparative evaluation and demonstrations is conducted in this research.
\par First, we employ different models suitable for MIMO, non-linear data. The training dataset is constructed by simulating various yaw and pitch angles ($\alpha$, $\beta$) in a physics-based environment and recording the corresponding tendon length variations ($L_1$, $L_2$, $L_3$). Each model has been trained and evaluated based on two key performance metrics: accuracy (MSE and MAE) and computation time.
\par Based on the experimental result, ensemble learning and kernel-based models both display significant score in MAE with medium computation time. Neural network model also shows promising results with high accuracy, stability, high learning efficiency, despite they are computationally intensive. Regularization-based model, on the other hand, requires less computational cost, making them efficient alternatives for real-time applications. Furthermore, this research utilize mathematical methods to derive control algorithms from each learning models, as a demonstration of transfer function identification. Eventually, we fed the enhanced function of GB model, which presents the best training result, back to the control system and conduct several robotic operations afterwards. The kinematic performance after the training process indicated that the robot can be manipulated more precisely, demonstrating the effectiveness of the proposed method.
\par In summary, this research aims to provide an well-structured framework for model-free control of bionic robots, and the findings can further contribute to various human-computer interaction (HCI) and robotics fields such as digital fabrication and biomedical industries in the long run.

\bibliographystyle{unsrtnat}
\bibliography{references}  






\end{document}